\documentclass[lettersize,journal]{IEEEtran}
\usepackage{amsmath,amsfonts}
\usepackage{array}
\usepackage{textcomp}
\usepackage{stfloats}
\usepackage{url}
\usepackage{verbatim}
\usepackage{cite}
\usepackage{xspace}
\usepackage{graphicx}
\usepackage{epstopdf}
\usepackage{amsmath,amssymb,booktabs,tabularx}
\usepackage{multirow}
\usepackage{algorithm}
\usepackage{algorithmicx}
\usepackage{algpseudocode}
\usepackage{bm}
\usepackage{subcaption}
\usepackage{xcolor}
\usepackage[pagebackref=true,breaklinks=true]{hyperref}
\hypersetup{
     colorlinks,
     linkcolor={red!75!black},
     citecolor={blue!75!black},
     urlcolor={blue!75!black},
}
\usepackage[capitalise]{cleveref}
\usepackage{microtype}
\usepackage{siunitx}
\usepackage{soul}
\usepackage{nicematrix}
\usepackage{orcidlink}

\makeatletter
\DeclareRobustCommand\onedot{\futurelet\@let@token\@onedot}
\newcommand{\@onedot}{\ifx\@let@token.\else.\null\fi\xspace}
\makeatother

\newcommand{\aka}{a.\,k.\,a\onedot}
\newcommand{\wrt}{w.\,r.\,t\onedot}

\newcommand{\ie}{i.\,e.,\xspace}
\newcommand{\eg}{e.\,g.,\xspace}
\newcommand{\vs}{vs\onedot}

\allowdisplaybreaks

\begin{document}
\makeatletter
\let\old@ps@headings\ps@headings
\let\old@ps@IEEEtitlepagestyle\ps@IEEEtitlepagestyle
\def\confheader#1{%
	\def\ps@IEEEtitlepagestyle{%
		\old@ps@IEEEtitlepagestyle%
		\def\@oddhead{\strut\hfill#1\hfill\strut}%
		\def\@evenhead{\strut\hfill#1\hfill\strut}%
	}%
	\ps@headings%
}
\makeatother

\confheader{
	\parbox{16cm}{ACCEPTED FOR PUBLICATION IN IEEE TRANSACTIONS ON GEOSCIENCE AND REMOTE SENSING.}
}
\title{AMD-HookNet++: Evolution of AMD-HookNet with Hybrid CNN-Transformer Feature Enhancement for Glacier Calving Front Segmentation}

\author{Fei Wu\orcidlink{0000-0003-4196-0289}, Marcel Dreier{\orcidlink{0009-0007-1929-1695}, Nora Gourmelon\orcidlink{0000-0003-3760-0184}, Sebastian Wind\orcidlink{0009-0004-7859-5785}, Jianlin Zhang\orcidlink{0000-0002-5284-2942}}, Thorsten Seehaus\orcidlink{0000-0001-5055-8959},\\Matthias Braun\orcidlink{0000-0001-5169-1567}, Andreas Maier\orcidlink{0000-0002-9550-5284}, and Vincent Christlein\orcidlink{0000-0003-0455-3799}

\thanks{Fei Wu, Marcel Dreier, Nora Gourmelon, Sebastian Wind, Andreas Maier and Vincent Christlein are with the Pattern Recognition Lab,  Department of Computer Science, Friedrich-Alexander-Universität Erlangen-Nürnberg, 91058 Erlangen, Germany.}
\thanks{Jianlin Zhang is with the School of Electrical, Electronics and Communication Engineering, University of Chinese Academy of Sciences, Beijing 100049, China, and also with the State Key Laboratory of Optical Field Manipulation Science and Technology, the Key Laboratory of Optical Engineering, Institute of Optics and Electronics, Chinese Academy of Sciences, Chengdu 610209, China.}
\thanks{Thorsten Seehaus and Matthias Braun are with the Department of Geography and Geosciences, Friedrich-Alexander-Universität Erlangen-Nürnberg, 91058 Erlangen, Germany.}
\thanks{Corresponding author: Fei Wu (email: river.wu@fau.de)}
}

\maketitle

\begin{abstract}
The dynamics of glaciers and ice shelf fronts significantly impact the mass balance of ice sheets and coastal sea levels. To effectively monitor glacier conditions, it is crucial to consistently estimate positional shifts of glacier calving fronts. However, laborious manual mapping calving fronts in satellite observations requires a considerable expense. The Attention-Multi-hooking-Deep-supervision HookNet (AMD-HookNet) firstly introduces a pure two-branch convolutional neural network (CNN) for glacier segmentation. Yet, the local nature and translational invariance of convolution operations, while beneficial for capturing low-level details, restricts the model ability to maintain long-range dependencies. In this study, we propose AMD-HookNet++, a novel advanced hybrid CNN-Transformer feature enhancement method for segmenting glaciers and delineating calving fronts in synthetic aperture radar (SAR) images. Our hybrid structure consists of two branches: a Transformer-based low-resolution (context) branch to capture long-range dependencies, which provides global contextual information in a larger view, and a CNN-based high-resolution (target) branch to preserve local details. To strengthen the representation of the connected hybrid features, we devise an enhanced spatial-channel attention (ESCA) module to foster interactions between the hybrid CNN-Transformer branches through dynamically adjusting the token relationships from both spatial and channel perspectives. Additionally, we develop a pixel-to-pixel contrastive deep supervision for optimizing our hybrid model. It integrates pixel-wise metric learning into glacier segmentation by guiding hierarchical pyramid-based pixel embeddings with category-discriminative capability. Through extensive experiments and comprehensive quantitative and qualitative analyses on the challenging glacier segmentation benchmark dataset CaFFe, we demonstrate that AMD-HookNet++ sets a new state of the art with an intersection over union (IoU) of 78.2 and a $95^\text{th}$ percentile Hausdorff distance (HD95) of 1,318\,m, while maintaining a competitive mean distance error (MDE) of 367\,m. More importantly, our hybrid model produces smoother delineations of calving fronts, resolving the issue of jagged edges typically seen in pure Transformer-based approaches.
\end{abstract}


\section{Introduction}
\IEEEPARstart{O}{ver} the past two decades, the glacier mass loss of polar ice sheets and other glaciated areas has accelerated increasingly~\cite{epic37530,tc-6-211-2012}. Changes to glacier terminus position, \ie glacier calving fronts, directly affect the glacier mass balance and, thus, contributing to global sea level rise and threatening coastal populations~\cite{frederikse2020causes,epic37530,Zemp2025}. Moreover, mass loss at marine-terminating outlet glaciers alters the forces governing glacier ice flow~\cite{liu2021automated,howat2008synchronous}, while buttressing forces diminish from glacier terminus leads to accelerated thinning of the glaciers, further exacerbating their dynamic mass loss~\cite{doi:10.1126/science.aae0017,https://doi.org/10.1029/2007JF000927,doi:10.1126/sciadv.aau8507}. Therefore, it is essential to continuously and accurately monitor the advance and retreat of glacier calving fronts to improve our knowledge of glaciological environments, \eg ice sheet mass balance, glacier dynamics, and their impact on sea level changes~\cite{vaughan2014observations,imbie2018mass,hugonnet2021accelerated,frank2022geometric,tc-19-5337-2025}. The growing availability of satellite observational data and technology plays an important role in monitoring and mapping glacier calving front locations. However, large-scale manual delineation of calving fronts is a costly, labor-intensive and time-consuming task. Annotators are susceptible to subjective errors when labeling marine-terminating outlet glaciers, which typically contain spectral surfaces similar to ice-melange (a mixture of sea ice and calved-off icebergs) in the vicinity of their calving fronts~\cite{marochov2021image}.

Recently, the developments of deep learning techniques in image segmentation~\cite{baumhoer2019automated,zhang2019automatically,zhang2021automated,mohajerani2019detection,heidler2021hed,liu2021multiscale,periyasamy2022get,holzmann2021glacier,baumhoer2021environmental,dong2022automatic,cheng2021calving,AMD-HookNet,10285497,10440599,essd-14-4287-2022,tc-2023-34,9563069,tc-18-3315-2024,ZHU2024609} have yielded significant advancements in the automated identification of glacier calving fronts. Most of these approaches~\cite{baumhoer2019automated,mohajerani2019detection,zhang2019automatically,liu2021multiscale,periyasamy2022get,holzmann2021glacier,AMD-HookNet,tc-2023-34,essd-14-4287-2022,heidler2021hed,9563069,loebel2022extracting,tc-18-3315-2024} build upon the traditional U-Net architecture~\cite{ronneberger2015u}, which is a convolutional neural network (CNN) based framework characterized by a symmetric U-shaped encoder-decoder structure with skip connections bridging the two structures. Instead of the typical single-branch-based U-Net methods~\cite{baumhoer2019automated,mohajerani2019detection,zhang2019automatically,liu2021multiscale,periyasamy2022get,holzmann2021glacier,essd-14-4287-2022,heidler2021hed,tc-2023-34,9563069,loebel2022extracting,tc-18-3315-2024}, AMD-HookNet~\cite{AMD-HookNet} first introduces two-branch U-Nets to glacier segmentation. One branch comprises a coarse resolution image used to provide global context, hence named the context branch. While the other branch comprises a center-cropped fine-grained resolution image used to capture local details, hence named the target branch. The two-branch contextual information interaction is realized by means of the spatial self-attention mechanism built upon the combined feature maps. AMD-HookNet~\cite{AMD-HookNet} shows a competitive glacier segmentation and calving front delineation performance on the challenging glacier segmentation benchmark dataset CaFFe~\cite{essd-14-4287-2022}. Motivated by the success of AMD-HookNet~\cite{AMD-HookNet}, HookFormer~\cite{10440599} explores the power of Vision Transformer (ViT) for glacier segmentation. It  is a pure two-branch ViT-based architecture which equipped with the Cross-Attention Swin-Transformer block within the Cross-Interaction module to model long-range dependencies and enhance multi-scale feature representation. So far, HookFormer~\cite{10440599} is the state-of-the-art glacier segmentation algorithm~\cite{gourmelon2025comparison} validated on the CaFFe dataset~\cite{essd-14-4287-2022}. However, as shown in \cref{figM0,fig07}, it can be observed that the ViT-based segmentation approaches, \eg Swin-Unet~\cite{cao2021swin}, MISSFormer~\cite{9994763}, and HookFormer~\cite{10440599} are easily to produce jagged edges for the detected calving fronts on the CaFFe dataset~\cite{essd-14-4287-2022}, which inevitably hinders the observation and understanding of glacier conditions. We argue that this phenomenon is caused by the limited data volume of the CaFFe dataset (681 images)~\cite{essd-14-4287-2022}, which is much smaller compared to the large-scale datasets, \eg ImageNet (1.28 million images)~\cite{deng2009imagenet}. With similar model capacity, ViT requires significantly more training data~\cite{wu2021cvt} than CNN to perform better due to the weak inductive bias of Transformer~\cite{pmlr-v238-zimerman24a}. An additional contributing factor may be that ViT lacks some of the inherent architectural advantages of CNNs, \eg the locality, the translation invariance, and the hierarchical structure, which are essential for visual perception tasks~\cite{10.1007/978-3-031-19806-9_13}. 

To alleviate this issue, some semantic segmentation algorithms try to introduce convolutions to ViTs~\cite{10281828,chen2024transunet,10285497,Maslov2025}. These hybrid CNN-Transformer models benefit from both architectural paradigms, with attention layers modeling long-range dependencies while convolutions emphasizing the local properties of images. However, their integration of local details and long-term correlations are realized by mixing CNN blocks and Transformer blocks within a single-branch-based framework. The single-branch-based segmentation framework has an upper limit of capturing long-range dependencies~\cite{Wu2025} (\ie global contextual information) since the long-distance dependencies are modeled by the large receptive fields formed by deep stacks of convolutional operations in CNN~\cite{8578911} or by the tokens in a sequence via self-attention in Transformer~\cite{vaswani2017attention}. While in these single-branch-based networks~\cite{Maslov2025,10285497,10281828,chen2024transunet}, all the hybrid operations are conducted on the shared feature space without independent task assignment. As a result, we believe that it is critical to develop a two-branch-based hybrid CNN-Transformer architecture to effectively leverage their respective strengths with distinct purposes in each branch, resulting in the proposed AMD-HookNet++. It is a novel advanced hybrid CNN-Transformer feature enhancement method for segmenting glaciers and delineating calving fronts in synthetic aperture radar (SAR) images. In particular, two branches designed with distinct purposes are involved in our AMD-HookNet++: the Transformer, renowned for its capacity to model long-range dependencies, is employed as the low-resolution context branch to capture and provide global contextual information. Concurrently, the CNN, which excels in refining local spatial details, is utilized as the high-resolution target branch to output the final prediction of the hybrid model. This dual-branch design leverages the complementary strengths of both architectures, thus ensuring a robust and detailed representation for the task at hand. To foster interactions between the hybrid CNN-Transformer branches, we devise an enhanced spatial-channel attention (ESCA) module to dynamically adjust the token relationships from both spatial and channel perspectives. Additionally, we develop a pixel-to-pixel contrastive deep supervision for optimizing our hybrid model. It integrates pixel-wise metric learning into glacier segmentation by guiding hierarchical pyramid-based pixel embeddings with category-discriminative capability. Through extensive experiments and comprehensive quantitative and qualitative analyses on the challenging glacier segmentation benchmark dataset CaFFe, we show that AMD-HookNet++ sets a new state of the art with an intersection over union (IoU) of 78.2 in glacier segmentation, outperforming the benchmark baseline: CaFFe~\cite{essd-14-4287-2022}, the base model: AMD-HookNet~\cite{AMD-HookNet}, and the state-of-the-art model: HookFormer~\cite{10440599} by the following absolute gains of \SI{8.5}{\percent}, \SI{3.8}{\percent}, and \SI{2.7}{\percent}, respectively. At the same time, AMD-HookNet++ achieves competitive performance in glacier calving front delineation with a mean distance error (MDE) of 367\,m and a $95^\text{th}$ percentile Hausdorff distance (HD95) of 1,318\,m, which is on par with the state-of-the-art model: HookFormer~\cite{10440599} while outperforming the benchmark baseline: CaFFe~\cite{essd-14-4287-2022} and the base model: AMD-HookNet~\cite{AMD-HookNet} by the following absolute gains of \SI{51.3}{\percent} and \SI{16.2}{\percent} on MDE, and \SI{39.5}{\percent} and \SI{19.2}{\percent} on HD95, respectively. More importantly, our hybrid model produces smoother delineations of calving fronts, resolving the issue of jagged edges typically seen in pure Transformer-based approaches~\cite{cao2021swin,9994763,10440599}. 

In summary, the main contributions of this work are as follows:
\begin{itemize}
    \item To mitigate the jagged artifacts typically seen in pure ViT-based approaches, we propose a novel hybrid CNN-Transformer architecture, termed AMD-HookNet++, for glacier segmentation and calving front delineation.
    \item We design an ESCA module aiming to robustly and efficiently fuse the hybrid CNN-Transformer features from both spatial and channel perspectives. 
    \item We develop a pixel-to-pixel contrastive deep supervision to optimize our hybrid model by integrating pixel-wise metric learning into glacier segmentation.
    \item We introduce the Hausdorff distance as a complementary metric to MDE to provide a more comprehensive assessment of calving front delineation. Our proposed method achieves state-of-the-art glacier segmentation performance on the CaFFe benchmark while maintaining competitive calving front delineation accuracy. We conduct extensive algorithm comparisons and in-depth component analyses to highlight its robustness and effectiveness.
\end{itemize}

The remainder of this paper is structured as follows: \cref{sec:related_work} provides an overview of the related deep learning works on glacier segmentation and semantic segmentation. The details of three core components defining our proposed method are described in \cref{sec:methodology}. \cref{sec:evaluation} reports the evaluation results, including the experimental setup, ablation studies, algorithm comparisons and analyses on the CaFFe dataset~\cite{essd-14-4287-2022}, along with a discussion of limitations and our future works. Finally, in \cref{sec:conclusion}, are drawn the conclusions of this article.

\section{Related Work}\label{sec:related_work}
\subsection{CNN-based glacier segmentation}
The current CNN-based methods for extracting glacier and ice-shelf fronts mainly rely on modifications to the popular U-Net framework~\cite{ronneberger2015u}, which consists of an encoder designed to capture semantic information and a symmetric reverse decoder, the latter of which allows for accurate pixel-level prediction while restoring image resolution. For instance, Baumhoer et al.~\cite{baumhoer2019automated} develop a modified U-Net to perform glacier segmentation on Sentinel-1 SAR imagery. This model is further used for analyzing potential relationships between calving front retreat and relative changes in the Antarctic environmental variables~\cite{baumhoer2021environmental}. Meanwhile, Mohajerani et al.~\cite{mohajerani2019detection} endeavor to adjust the configuration of U-Net to trace glacier fronts in low-resolution optical Landsat data while Zhang et al~\cite{zhang2019automatically} apply a large-kernel-based U-Net to delineate glacier calving fronts in high-resolution TerraSAR-X images. Later, Liu et al.~\cite{liu2021multiscale} employ a multi-scale U-Net to extract glacier contours in single-polarimetric SAR images, whereas Periyasamy et al.~\cite{periyasamy2022get} examine how varying hyper-parameter configurations affect glacier segmentation during the U-Net training. Rather than solely investigating the U-Net architecture, Holzmann et al.~\cite{holzmann2021glacier} enhance the standard skip connections of the U-Net by including attention gates to preserve critical information. Notably, Heidler et al.~\cite{heidler2021hed} showcase significant advancements in glacier segmentation and calving front detection using a U-Net-based multi-task learning framework, which inspires Herrmann et al.~\cite{tc-2023-34} to delve into a multi-task learning strategy within the nnU-Net architecture~\cite{Isensee2021} for glacier front segmentation. As discussed above, the U-Net-based approaches~\cite{baumhoer2019automated,heidler2021hed,holzmann2021glacier,mohajerani2019detection,liu2021multiscale,zhang2019automatically,periyasamy2022get,AMD-HookNet,tc-2023-34,essd-14-4287-2022} are predominant in glacier segmentation task. While binary segmentation is commonly employed~\cite{baumhoer2019automated,mohajerani2019detection,heidler2021hed,holzmann2021glacier,zhang2019automatically,periyasamy2022get,cheng2021calving}, some works~\cite{marochov2021image,AMD-HookNet,essd-14-4287-2022,tc-2023-34} extend this to multi-class (zone) segmentation. Directly delineating glacier calving fronts is often error-prone~\cite{heidler2021hed} owing to severe class imbalance issues~\cite{fidon2017generalised}. As a result, most research~\cite{baumhoer2019automated,periyasamy2022get,holzmann2021glacier,mohajerani2019detection,liu2021multiscale,zhang2019automatically,AMD-HookNet,tc-2023-34} rely on post-processing techniques to derive calving fronts from the multi-class segmentation outputs. This pipeline is also followed in the present study.

In contrast to the aforementioned adaptations of U-Net~\cite{ronneberger2015u}, Selbesoğlu et al.~\cite{drones7020072} conduct a comprehensive analysis and evaluation of three state-of-the-art deep learning architectures~\cite{chen2017rethinking,NEURIPS2021_64f1f27b,NEURIPS2021_55a7cf9c} for the segmentation of glaciers in the Antarctic Peninsula. Dong et al.~\cite{dong2022automatic} develop a game theory-based framework for distinguishing between outlet glaciers and ocean using available Digital Elevation Model (DEM) products while Marochov et al.~\cite{marochov2021image} introduce a two-stage methodology rooted in the VGG architecture~\cite{simonyan2014very} to achieve granular pixel-level classification across glacial terrains. Furthermore, three studies~\cite{cheng2021calving,zhang2021automated} advance the application of the DeepLabV3 network~\cite{chen2017rethinking} for glacier segmentation. Cheng et al.~\cite{cheng2021calving} refine both pre- and post-processing tools with a modified DeepLabV3+Xception model~\cite{10.1007/978-3-030-01234-2_49}, facilitating the monitoring of tidewater glacier front dynamics in Greenland. Concurrently, Zhang et al.~\cite{zhang2021automated} investigate the integration of DeepLabV3 with multiple established backbone networks~\cite{he2016deep,yu2017dilated,howard2017mobilenets} to enhance segmentation performance. They later introduce an automated pipeline AutoTerm~\cite{tc-17-3485-2023} using the DeepLabV3+ architecture~\cite{10.1007/978-3-030-01234-2_49} for extracting glacier terminus traces from multi-source remotely sensed data in Greenland.

Our previously proposed AMD-HookNet~\cite{AMD-HookNet} integrates dual U-Net~\cite{ronneberger2015u} branches operating on heterogeneous spatial resolutions to address glacier front segmentation in SAR imagery. This multi-scale CNN framework enables hierarchical integration of auxiliary contextual data from the low-resolution pathway (global context branch) into the high-resolution pathway (local target branch), enhancing feature representation. A self-attention mechanism embedded within the feature alignment (\emph{hooking}) makes the network focus on learning critical spatial dependencies. Moreover, an iterative multi-hooking mechanism coupled with deep supervision further optimizes the training process, yielding substantial performance improvement in both glacier segmentation and calving front delineation.

\subsection{ViT-based semantic segmentation}
Transformer was initially introduced in natural language processing (NLP)~\cite{vaswani2017attention} but later extended to various computer vision tasks due to its powerful global modeling capabilities, \aka Vision Transformers (ViTs). Many works investigate ViT into semantic segmentation domain~\cite{9994763,liu2021swin,cao2021swin,10440599,wang2021crossformer,yang2021focal}. For example, Swin-Transformer~\cite{liu2021swin} is a prominent ViT backbone which employs shifted-window self-attention to obtain hierarchical feature representations. Building upon this architecture, Swin-Unet~\cite{cao2021swin} develops a symmetric U-shaped encoder-decoder structure for medical image segmentation. MISSFormer~\cite{9994763} introduces a hierarchical U-shaped Transformer which coupled with the Remixed Transformer Context Bridge module to extract multi-scale discriminative features for medical image segmentation. In contrast, CrossFormer~\cite{wang2021crossformer} integrates a cross-scale embedding layer with a long-short distance attention mechanism to facilitate cross-attention, which is implemented on the concatenated attention maps derived from multi-scale tokens within a single input image. Focal-Transformer~\cite{yang2021focal}, on the other hand, introduces focal self-attention into ViT by aggregating multi-granularity features from surrounding regions using progressively squeezed window sizes, thus maintaining a broad global perspective. More recently, HookFormer~\cite{10440599} establishes a pure two-branch-based multi-scale Transformer architecture for glacier segmentation. The global-local tokens, derived from the context and target branches, are exclusively processed through the Cross-Attention Swin-Transformer block within the Cross-Interaction module for global context modeling and fine-grained feature interaction. As a result, HookFormer~\cite{10440599} achieves the state-of-the-art glacier segmentation and calving front delineation performance on the CaFFe dataset~\cite{essd-14-4287-2022}.

\subsection{Hybrid CNN-Transformer-based semantic segmentation}
Hybrid models combining CNNs and Transformers have yielded many successful applications in semantic segmentation~\cite{chen2024transunet,9706678,10.1007/978-3-031-08999-2_22,Maslov2025,10285497,10281828,9878483,NEURIPS2021_950a4152,10244199,lee2022mpvit}. As a representative hybrid CNN-Transformer model for 2D medical image segmentation, Trans-UNet~\cite{chen2024transunet} adds Transformer layers in the deepest part of the CNN encoder to capture global contextual information. Conversely, UNETR~\cite{9706678} and Swin UNETR~\cite{10.1007/978-3-031-08999-2_22} adopt Transformers in their encoder while using a convolutional decoder for 3D medical image segmentation. Based on a multi-scale patch embedding and multi-path structure scheme, Lee et al.~\cite{lee2022mpvit} concatenate the convolutional local features to the Transformer's global features for global-to-local interaction. In the glacier segmentation domain, GLA-STDeepLab~\cite{10285497} is a pioneering hybrid CNN-Transformer effort that incorporates the Swin-Transformer block~\cite{liu2021swin} into DeepLabV3+ architecture~\cite{10.1007/978-3-030-01234-2_49}  to empower the long-range contextual dependency modeling. Specifically, it inserts Transformer blocks into the top of (the bottom of) the DeepLabV3+ encoder (decoder) for the purpose of processing partitioned tokens (recovering the image resolution). Instead of mixing Transformer and CNN blocks simultaneously,  GlaViTU~\cite{10281828} consecutively applies the SEgmentation TRansformer (SETR)~\cite{9578646} followed by a modified U-Net~\cite{ronneberger2015u} to form a hybrid model for multi-regional glacier mapping. This method is later employed to track glacier changes on a multi-modal global-scale remote sensing dataset~\cite{Maslov2025}.
\begin{figure*}[!t]
\centering
\includegraphics[width=.99\textwidth]{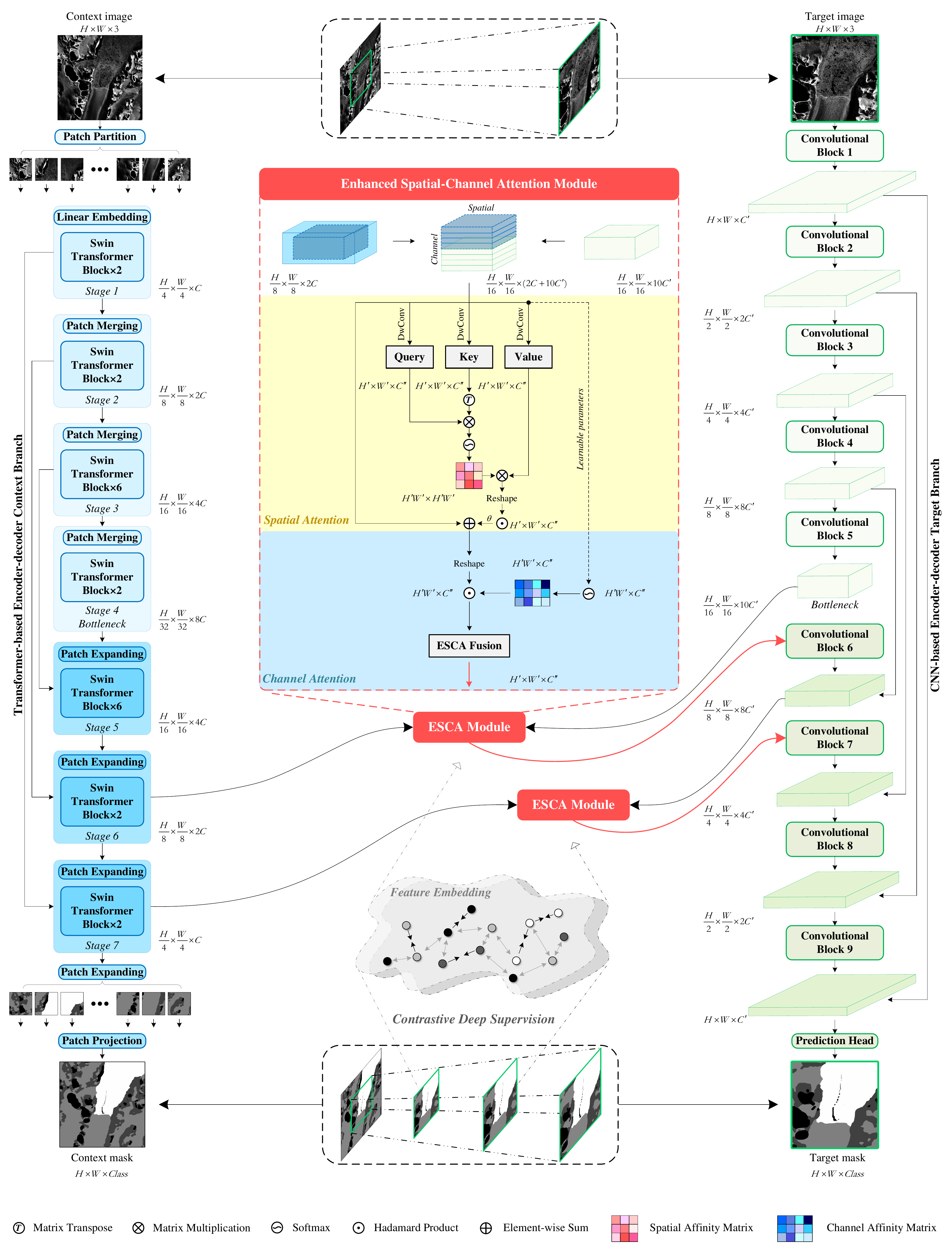}
\caption{Overview of our hybrid CNN-Transformer-based AMD-HookNet++ approach.}\label{fig00}
\end{figure*}

\section{Methodology}\label{sec:methodology}
Our AMD-HookNet~\cite{AMD-HookNet} utilizes a multi-scale CNN architecture capable of exchanging fine- and coarse-grained information and exhibiting sustained performance gains in glacier segmentation and calving front delineation. We later introduce a pure ViT framework HookFormer~\cite{10440599} to fully explore the capabilities of modeling long-range contextual dependencies in Transformers. However, we observe severe jagged artifacts for the detected calving fronts in HookFormer~\cite{10440599}, as well as other ViT-based segmentation approaches, \eg Swin-Unet~\cite{cao2021swin} and MISSFormer~\cite{9994763} (see \cref{figM0,fig07}). This inevitably hinders the observation and understanding of glacier conditions. CNN is good at extracting local details, but it suffers from insufficient receptive field. While Transformer has the advantage of global representation, it ignores local information especially when trained on comparatively small datasets~\cite{wu2021cvt}. Considering the balance of localized spatial details in CNN and global relationships in Transformer, we now propose AMD-HookNet++, the evolution of AMD-HookNet~\cite{AMD-HookNet} with hybrid CNN-Transformer feature enhancement for glacier calving front segmentation. 

The advancements in AMD-HookNet++ can be categorized into three key aspects: (1)~the novel hybrid two-branch structure (\cref{NA}): a Transformer-based low-resolution (context) branch to capture long-range dependencies, which provides global contextual information within a larger view to a CNN-based high-resolution (target) branch to refine local fine-grained details, (2)~the enhanced spatial-channel attention module (\cref{ESCA}) to foster interactions between the hybrid CNN-Transformer branches, and (3)~the integration of the developed pixel-to-pixel contrastive deep supervision (\cref{CDS}) for optimizing our hybrid model.

\subsection{Network Architecture}
\label{NA}
The existing hybrid CNN-Transformer segmentation approaches either customize hybrid CNN-Transformer blocks within the network~\cite{chen2024transunet,10285497}, or use consecutive hierarchical Transformer blocks to replace the CNN encoder~\cite{9706678,10.1007/978-3-031-08999-2_22} or decoder~\cite{9878483,NEURIPS2021_950a4152} of the U-shaped structure. However, all the hybrid operations in these single-branch-based works are conducted on the shared feature space without independent task assignment. This leads to a constrained global perspective and inadequate long-range dependencies, \ie no more additional information can be introduced except for the visual content in the same original input image. 

Therefore, we propose a novel advanced hybrid CNN-Transformer feature enhancement method (AMD-HookNet++) for glacier segmentation and calving front delineation. \Cref{fig00} illustrates the architectural details. As an evolutionary work, the model design of AMD-HookNet++ adheres to the two-branch paradigm established by AMD-HookNet~\cite{AMD-HookNet}, which comprises a context branch and a target branch. Nonetheless, to fully take advantage of the global modeling capabilities of Transformer, the context branch for capturing long-range dependencies from coarse-grained low-resolution image is built upon a pure ViT, specifically Swin-Unet~\cite{cao2021swin}. In contrast, the target branch which focuses on extracting localized spatial details from fine-grained high-resolution image is built upon a pure CNN. Note that the context branch is intended to provide global contextual information to the target branch, \ie there is a semantic connection between the two input images from the context branch and the target branch. Concretely, the image resolutions for the target branch ($\mathrm{r_t}$) and the context branch ($\mathrm{r_c}$) conform to the matching paradigm as:
\begin{equation}
     \mathrm{r_{t}}=2\times\mathrm{r_{c}}
\label{equ_1}
\end{equation}

For the same region, the resolution of the target branch ($\mathrm{t}$) is twice that of the context branch ($\mathrm{c}$). This process is implemented by center-cropping the target input from the original context input, while this original context input is subsequently downsampled to match the input size of the target branch. In this way, the context branch ($\mathrm{c}$) not only preserves the same semantic content as the target branch but also incorporates broader contextual surroundings with reduced high-frequency detail. This contributes to the extraction of global contextual information around the localized target region. The auxiliary function of context enhancement is realized by aligning feature maps from the context branch to the target branch, \aka \emph{hooking}. Note that we replace the self-attention-based \emph{hooking} (channel-wise feature concatenation followed by self-attention) of AMD-HookNet~\cite{AMD-HookNet} with our enhanced spatial-channel attention (ESCA) module, which is designed to enable complementary guidance and effective interaction of the hybrid model, \ie efficient information exchange between the Transformer and the CNN feature maps. Please refer to \cref{ESCA} for more details.
\begin{figure}[!t]
\centering
\includegraphics[width=.35\textwidth]{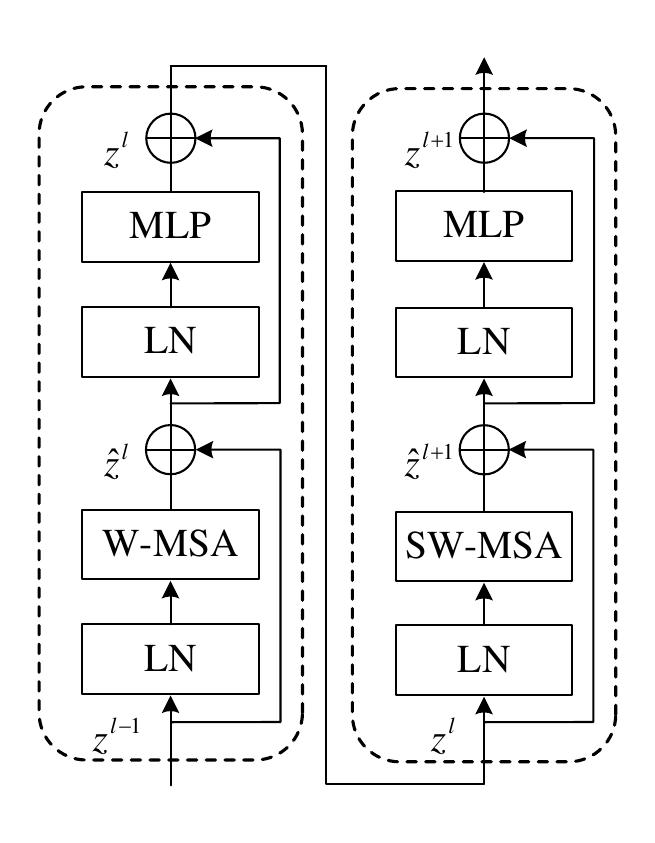}
\caption{Two successive Swin-Transformer blocks~\cite{liu2021swin}.}
\label{fig01}
\end{figure}
\begin{table*}[t]
	\centering
	\caption{The two-branch layer specifications of AMD-HookNet++.}
 \begin{tabular*}{\textwidth}{@{\extracolsep{\fill}}llclc@{\extracolsep{\fill}}}
	\toprule
        &\multicolumn{2}{c}{Transformer-based context branch}&\multicolumn{2}{c}{CNN-based target branch} \\
        &Layer specification&Feature size&Layer specification&Feature size \\ 
      \midrule
	&Input&\numproduct{224x224x3}&Input&\numproduct{224x224x3} \\
        \midrule
        \multirow{5}{*}{Encoder}&Patch Embedding&\numproduct{56x56x96}&Convolution Block 1&\numproduct{224x224x32} \\
        &Stage 1&\numproduct{56x56x96}&Convolution Block 2&\numproduct{112x112x64} \\
        &Stage 2&\numproduct{28x28x192}&Convolution Block 3&\numproduct{56x56x128} \\
        &Stage 3&\numproduct{14x14x384}&Convolution Block 4&\numproduct{28x28x256} \\
        &Stage 4&\numproduct{7x7x768}&Convolution Block 5&\numproduct{14x14x320} \\
        \midrule
        \multirow{5}{*}{Decoder}&Stage 5&\numproduct{14x14x384}&Convolution Block 6&\numproduct{28x28x256} \\
        &Stage 6&\numproduct{28x28x192}&Convolution Block 7&\numproduct{56x56x128} \\
        &Stage 7&\numproduct{56x56x96}&Convolution Block 8&\numproduct{112x112x64} \\
        &Patch Expanding&\numproduct{224x224x96}&Convolution Block 9&\numproduct{224x224x32} \\
        &Patch Projection&\numproduct{224x224x4}&Prediction Head&\numproduct{224x224x4} \\
        \bottomrule
 \end{tabular*}
\label{tab00}
\end{table*}

\subsubsection{Transformer-based context branch}
\label{TB}
The context branch of our AMD-HookNet++ adopts Swin-Unet~\cite{cao2021swin}, which is a pure ViT architecture equipped with symmetric U-shaped Swin-Transformer~\cite{liu2021swin}. By leveraging the hierarchical multi-scale representation enabled through an efficient shifted window partitioning strategy, Swin-Transformer~\cite{liu2021swin} functions as a versatile ViT backbone and is adaptable to a wide range of computer vision applications~\cite{10440599,cao2021swin,10.1007/978-3-031-08999-2_22}. As depicted in \cref{fig01}, a pair of successive Swin-Transformer blocks is presented. Each block comprises Layer Normalization (LN), a multi-head self-attention module, a residual connection, and a two-layer perceptron (MLP) employing Gaussian Error Linear Unit (GELU) activation. Within consecutive Swin-Transformer blocks, the first block incorporates a window-based multi-head self-attention (W-MSA) module, while the subsequent block utilizes a shifted window-based multi-head self-attention (SW-MSA) module. The computational process for these sequential blocks is defined as follows:

\begin{align}
    \hat{z}^{l}&=\mathrm{W\mbox{-}MSA}(\mathrm{LN}(z^{l-1}))+z^{l-1}\\
    z^{l}&=\mathrm{MLP}(\mathrm{LN}(\hat{z}^{l}))+\hat{z}^{l}\\
    \hat{z}^{l+1}&=\mathrm{SW\mbox{-}MSA}(\mathrm{LN}(z^{l}))+z^{l}\\
    z^{l+1}&=\mathrm{MLP}(\mathrm{LN}(\hat{z}^{l+1}))+\hat{z}^{l+1}
\end{align}
where $\hat{z}^{l}$ and $\hat{z}^{l+1}$ represent the outputs of the residual connection following the W-MSA and SW-MSA modules in the $l^\text{th}$ and $(l+1)^\text{th}$ Swin-Transformer blocks, respectively. Meanwhile, $z^{l}$ and $z^{l+1}$ represent the outputs of the successive Swin-Transformer blocks. Self-attention, a fundamental mechanism in ViTs~\cite{vaswani2017attention,liu2021swin,9994763}, is formally defined as:
\begin{equation}
    \mathrm{Self\mbox{-}attention}(Q,K,V)=\mathrm{Softmax}\Big(\frac{Q\,K^\top}{\sqrt{d_{k}}}\Big)V
    \label{eq:sa}
\end{equation}
where $Q,K\in{\mathbb{R}^{N^{2}\times{d_{k}}}}$ and $V\in{\mathbb{R}^{N^{2}\times{d_{v}}}}$ are the query, key, and value matrices, respectively, obtained via independent linear transformations of the input tokens. $N^{2}$ signifies the cardinality of tokens within a given window while $d_{k}$ and $d_{v}$ represent the respective embedding dimensions of the query/key and value matrices. Notably, a relative positional bias term is incorporated into the $\mathrm{Softmax}$ output, encoding the spatial displacement between the two windows.

\subsubsection{CNN-based target branch}
\label{CB}
The Transformer-based low-resolution (context) branch provides long-range contextual dependencies within a global view, while a CNN-based high-resolution (target) branch is employed to capture localized spatial details. Our CNN-based target branch adopts a classic U-Net encoder-decoder design, which incorporates nine convolutional blocks for encoding/decoding and a single \numproduct{1x1} convolutional prediction head for pixel-wise multi-class categorization. Every convolutional block comprises two sequential units, each containing a \numproduct{3x3} convolutional layer with a padding size of 1, followed by Batch Normalization (BN), a Rectified Linear Unit (ReLU) activation function, and either a max-pooling layer (in the encoder) or a deconvolutional layer (in the decoder). The base number of channels per convolutional filter starts at 32 and increases proportionally with the network depth until reaching the bottleneck. The precise channel configurations for the two-branch structure of AMD-HookNet++ are detailed in \Cref{tab00}, where feature size corresponds to the spatial dimensions of the activation maps, \ie their volumetric resolution.

\subsection{Enhanced Spatial-Channel Attention Module}
\label{ESCA}
How to effectively combine CNN and Transformer features is an open question in the field~\cite{chen2024transunet,9706678,10.1007/978-3-031-08999-2_22,Maslov2025,10285497,10281828,9878483,NEURIPS2021_950a4152,10244199,lee2022mpvit}. Some of the existing hybrid CNN-Transformer models rely on element-wise addition~\cite{10244199} or simple feature concatenation~\cite{lee2022mpvit}. Attention mechanism serves a pivotal function in computer vision by emulating the human cognitive process of selectively focusing on salient information, while simultaneously disregarding non-essential data and suppressing redundant or extraneous signals~\cite{AMD-HookNet,vaswani2017attention}. AMD-HookNet~\cite{AMD-HookNet} introduces self-attention based on the concatenated two-branch CNN features to automatically adjust spatial relationships with learnable weights, thus revealing improved performance compared to the simple feature concatenation used in HookNet~\cite{van2021hooknet}. However, we contend that relying solely on single spatial self-attention for merging two-branch features is suboptimal, as it fails to account for the complementary information between channels and spatial positions. In deep neural networks (DNNs), the allocation of attention is typically governed by two distinct mechanisms: spatial attention determines the specific spatial locations to prioritize (where), while channel attention identifies the relevant feature channels to emphasize (what)~\cite{HUANG2024108784,Guo2022}. SENet~\cite{8701503} is a pioneering work in channel attention which introduces squeeze-and-excitation block with global average pooling to explicitly model the interdependencies among feature channels. Later, CBAM~\cite{10.1007/978-3-030-01234-2_1} sequentially combines channel attention and spatial attention with both global average pooling and global max pooling to aggregate features, enabling the network to selectively emphasize informative channels and spatially important regions. Nevertheless, global average pooling is easy to make the channel descriptors have homogeneity while global average pooling is prone to network overfitting~\cite{DBLP:journals/corr/abs-1301-3557}, which inevitably weakens the detailed distinction between feature maps~\cite{LUO2020119}. Moreover, the spatial attention map is computed by compressing channels, which results in a uniform distribution of spatial attention weights across all channels during element-wise multiplication with the input features. Consequently, the adaptive capacity of attention is restricted as the spatial attention weights cannot be dynamically recalibrated \wrt the specific characteristics of each channel~\cite{HUANG2024108784}. 
\begin{figure}[!t]
\centering
\includegraphics[width=.5\textwidth]{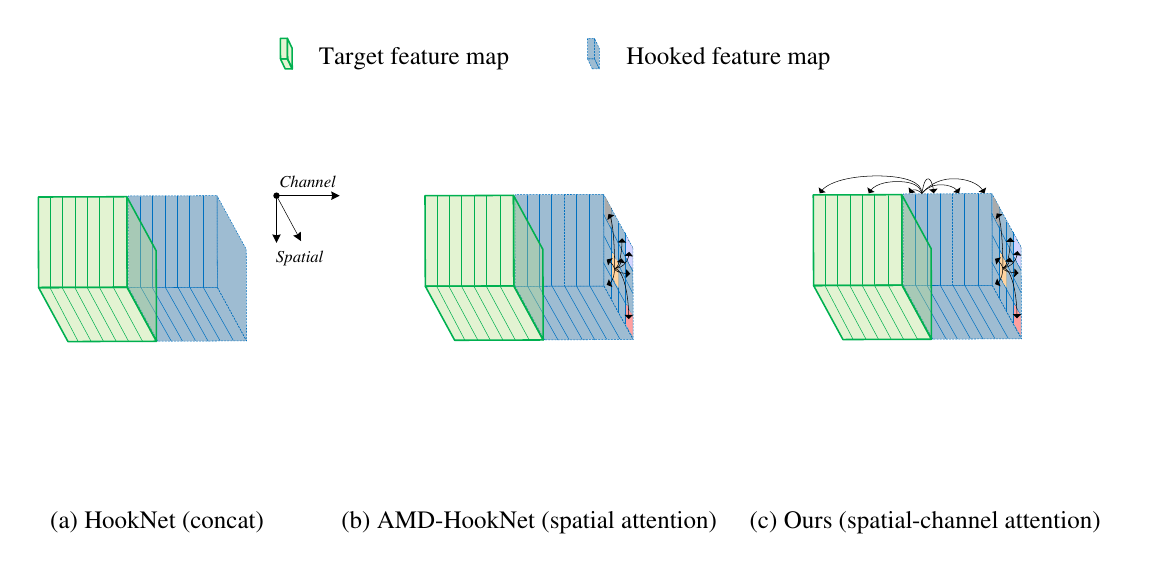}
\caption{Comparative analysis of two-branch information interaction strategies, including (a) feature concatenation used in HookNet~\cite{van2021hooknet}, (b) self-attention based on feature concatenation used in AMD-HookNet~\cite{AMD-HookNet}, and (c) spatial-channel attention between two-branch features used in our AMD-HookNet++.}
\label{fig02}
\end{figure}

\begin{figure*}[!t]
\centering
\includegraphics[width=\textwidth]{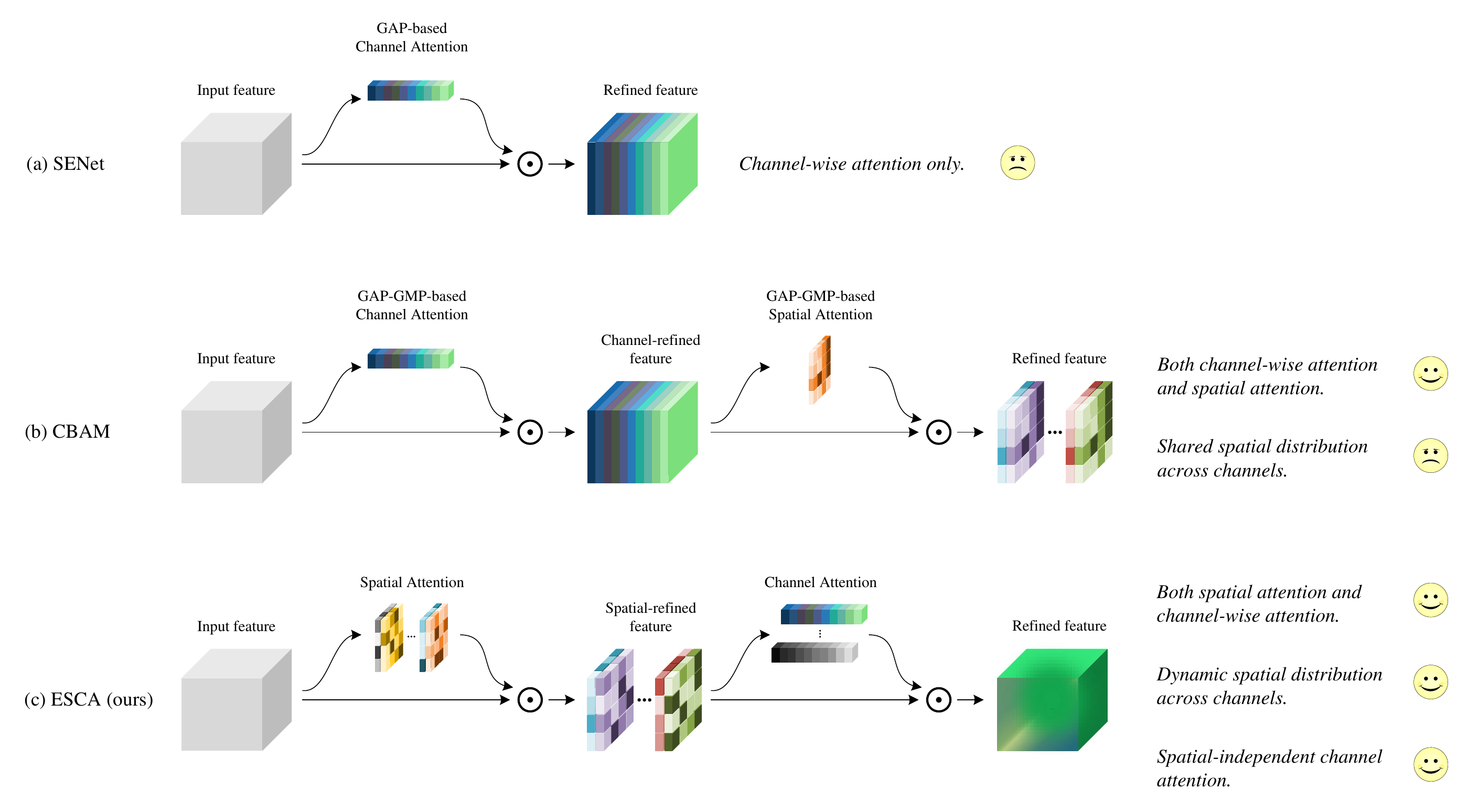}
\caption{Schematic representation of the refined features of the three attention mechanisms: (a) SENet~\cite{8701503}, (b) CBAM~\cite{10.1007/978-3-030-01234-2_1}, and (c) ESCA (ours), where GAP and GMP denote global average pooling and global max pooling, respectively.}\label{fig03}
\end{figure*}

As an evolutionary work of AMD-HookNet~\cite{AMD-HookNet}, this study focuses on exploring attention-guided strategies for fusing CNN-Transformer features, resulting in our enhanced spatial-channel attention (ESCA) module. \Cref{fig02} compares the difference in \emph{hooking} mechanisms among HookNet~\cite{van2021hooknet}, AMD-HookNet~\cite{AMD-HookNet}, and the proposed AMD-HookNet++. Unlike HookNet~\cite{van2021hooknet} and AMD-HookNet~\cite{AMD-HookNet}, which utilize feature concatenation and self-attention mechanism operating on concatenated features for \emph{hooking}, we devise the ESCA module for more effective global-local information interactions from both spatial and channel perspectives. In particular, \cref{fig03} illustrates the schematic representation of typical attention modules~\cite{8701503,10.1007/978-3-030-01234-2_1} and our ESCA module. We can see that SENet~\cite{8701503} exclusively employs channel attention, thereby lacking the capacity to prioritize significant spatial regions. As discussed above, although CBAM~\cite{10.1007/978-3-030-01234-2_1} includes both channel and spatial attention, it applies the same spatial attention pattern uniformly to all channels within its refined feature outputs, which may constrain its adaptability to channel-specific spatial information. More importantly, the squeeze operation (i.e., feature aggregation) is either realized by global average pooling in SENet~\cite{8701503} or by the combination of global average pooling and global max pooling in CBAM~\cite{10.1007/978-3-030-01234-2_1}. This process extracts the general feature representations alongside the channel/spatial dimension but is too simple to capture complex global information~\cite{Guo2022}. To prevent the loss of critical information, our proposed ESCA module differs from the conventional squeeze-and-excitation attention pipeline~\cite{8701503,10.1007/978-3-030-01234-2_1,Guo2022}. Instead of applying dimensional compression, the inputs for computing spatial and channel attention are derived directly from the original features. Building on this, the sequential structure of the ESCA module, processing spatial attention prior to channel attention, enables a dynamic and adaptive recalibration of attention weights across both spatial and channel dimensions. This design enhances the representational capacity of the attention mechanisms, offering a more robust and effective feature refinement. \Cref{fig00} depicts the structural specifics of our ESCA module.

\subsubsection{Spatial Attention}
\label{SA}
The two input feature maps of the ESCA module initialize the \emph{hooking} of the hybrid CNN-Transformer branches, where one feature map $F_{\mathrm{c}}\in\{\mathbb{R}^{\frac{H}{8}\times\frac{W}{8}\times2C},\mathbb{R}^{\frac{H}{4}\times\frac{W}{4}\times{C}}\}$ is derived from \textit{Stage 6} and \textit{Stage 7} of the Transformer-based context branch ($\mathrm{c}$) while the other feature map $F_{\mathrm{t}}\in\{\mathbb{R}^{\frac{H}{16}\times\frac{W}{16}\times{10C'}},\mathbb{R}^{\frac{H}{8}\times\frac{W}{8}\times{8C'}}\}$ is derived from \textit{Convolutional Block 5} and \textit{Convolutional Block 6} of the CNN-based target branch ($\mathrm{t}$). The input matrix $\bm{M}$ for calculating spatial attention is originated from the channel-wise concatenation between the center-cropped context feature $F_{\mathrm{c}}$ and the target feature $F_{\mathrm{t}}$, which can be expressed as:
\begin{equation}
\bm{M}=\mathrm{Concat}\left(\mathrm{Center\mbox{-}crop}(F_{\mathrm{c}}),F_{\mathrm{t}}\right)
\label{eq7}
\end{equation}
where the $H$ and $W$ correspond to the height and width of the input image dimensions, respectively. $C$ represents the number of channels in the tokenized input from the context branch, while $C'$ refers to the number of channels in the convoluted input from the target branch. To extract channel-independent matrix, we use depth-wise convolution to construct self-attention, where a distinct spatial attention map is computed independently for each channel. This can be formulated as:
\begin{equation}
    \bm{q}=\mathrm{DwConv}(\bm{M})
\end{equation}
\begin{equation}
    \bm{k}=\mathrm{DwConv}(\bm{M})
\end{equation}
\begin{equation}
    \bm{v}=\mathrm{DwConv}(\bm{M})
\end{equation}
\begin{equation}
    \mathrm{Self\mbox{-}attention}(\bm{q},\bm{k},\bm{v})=\mathrm{Softmax}\Big(\frac{\bm{q}\,\bm{k}^\top}{\sqrt{d}}\Big)\bm{v}
\end{equation}
where $\bm{q},\bm{k},\bm{v}\in{\mathbb{R}^{H'\times{W'}\times{C''}}}$ represent the query, key, and value matrices, respectively, generated through independent depth-wise convolutional layers applied to the input matrix $\bm{M}$. $d$ denotes the dimension of the query, key, or value matrices and is used to scale the function. $H'$, $W'$, and $C''$ correspond to the height, width, and channel count of the specific input matrix $\bm{M}$, respectively. Further, we integrate a learnable parameter $\theta$ alongside a skip connection within the spatial attention module:
\begin{multline}
    \mathrm{Spatial\mbox{-}attention}(\bm{q},\bm{k},\bm{v})=\\\bm{M}+\mathrm{\theta}\times\mathrm{Self\mbox{-}attention}(\bm{q},\bm{k},\bm{v})
\end{multline}
where $\theta$ is used to adjust the weights of the self-attention while the skip connection improves model convergence.

\subsubsection{Channel Attention}
\label{CA}
Each channel of a high-level feature represents a specific semantic response~\cite{Guo2022}. Instead of computing the channel attention map using the typical squeeze-and-excitation~\cite{8701503,10.1007/978-3-030-01234-2_1,Guo2022}, we specifically design a channel attention module to explore the inter-channel relationships presented within the spatial-attention-refined channel-independent features. Given the input matrix $\bm{M}\in{\mathbb{R}^{H'\times{W'}\times{C''}}}$, we detach a matrix $\bm{U}\in{\mathbb{R}^{H'W'\times{C''}}}$ with learnable parameters from $\bm{M}$. Followed by a $\mathrm{Softmax}$ function, this matrix $\bm{U}$ is used to encode the spatial-independent channel affinity $\bm{U'}$, which can be expressed as:
\begin{equation}
    \bm{U'}=\mathrm{Softmax}(\bm{U}, dim=C'')
\end{equation}
where $\bm{U'}\in{\mathbb{R}^{H'W'\times{C''}}}$. Afterward, we reshape the feature refined by spatial attention to match the dimensions of $\bm{U'}$. This reshaped feature is then multiplied by the spatial-independent channel affinity matrix $\bm{U'}$ to generate the channel attention map.
\begin{multline}
    \mathrm{Channel\mbox{-}attention}=\\\bm{U'}\times{\mathrm{Reshape}(\mathrm{Spatial\mbox{-}attention}(\bm{q},\bm{k},\bm{v}))}
\end{multline}

Ultimately, we utilize an ESCA fusion that includes a dimensional reshape operation followed by a point-wise convolution with shared parameters to maintain spatial diversity and produce the refined feature for the ESCA module, \ie as the \emph{hooking} mechanism between the hybrid CNN-Transformer branches.
\begin{equation}
    \mathrm{ESCA\mbox{-}hooking}=\mathrm{ESCA\mbox{-}fusion}(\mathrm{Channel\mbox{-}attention})
\end{equation}

\subsection{Contrastive Deep Supervision and Loss Function}
\label{CDS}
In traditional supervised learning, supervision is applied solely to the final output layer, with errors being propagated backward from the last layer to the earlier layers. This leads to difficulties in optimizing intermediate layers like gradient vanishing~\cite{10.1007/978-3-031-19809-0_1}. To this end, deep supervision was proposed to incorporate multiple auxiliary classifiers at various intermediate layers of the network to improve model convergence~\cite{lee2015deeply}. There already exist several successful implementations of deep supervision in glacier segmentation, such as AMD-HookNet~\cite{AMD-HookNet} and HookFormer~\cite{10440599}. However, Zhang et al.~\cite{10.1007/978-3-031-19809-0_1} point out that the shallow layers learn task-irrelevant low-level features which conflict with the task-biased loss functions, resulting in their contrastive deep supervision. By optimizing the intermediate layers with contrastive learning instead of traditional supervised learning, they show that using contrastive learning enhances the visual representation in image classification and object detection tasks. On the other hand, Zhou et. al~\cite{10443562} advocate a pixel-to-pixel contrastive learning as a new supervised learning paradigm for semantic segmentation. The design of contrastive learning for semantic segmentation is significantly more intricate compared to image classification or object detection. Image classification and object detection tasks involve assigning either a single class label to the entire image or a limited set of class labels to specific regions or objects, respectively. In contrast, semantic segmentation requires pixel-wise dense predictions, where each pixel must be assigned a specific class label, \ie the “region” involved in contrastive learning is reduced to a single pixel. 
\begin{figure}[!t]
\centering
\includegraphics[width=.45\textwidth]{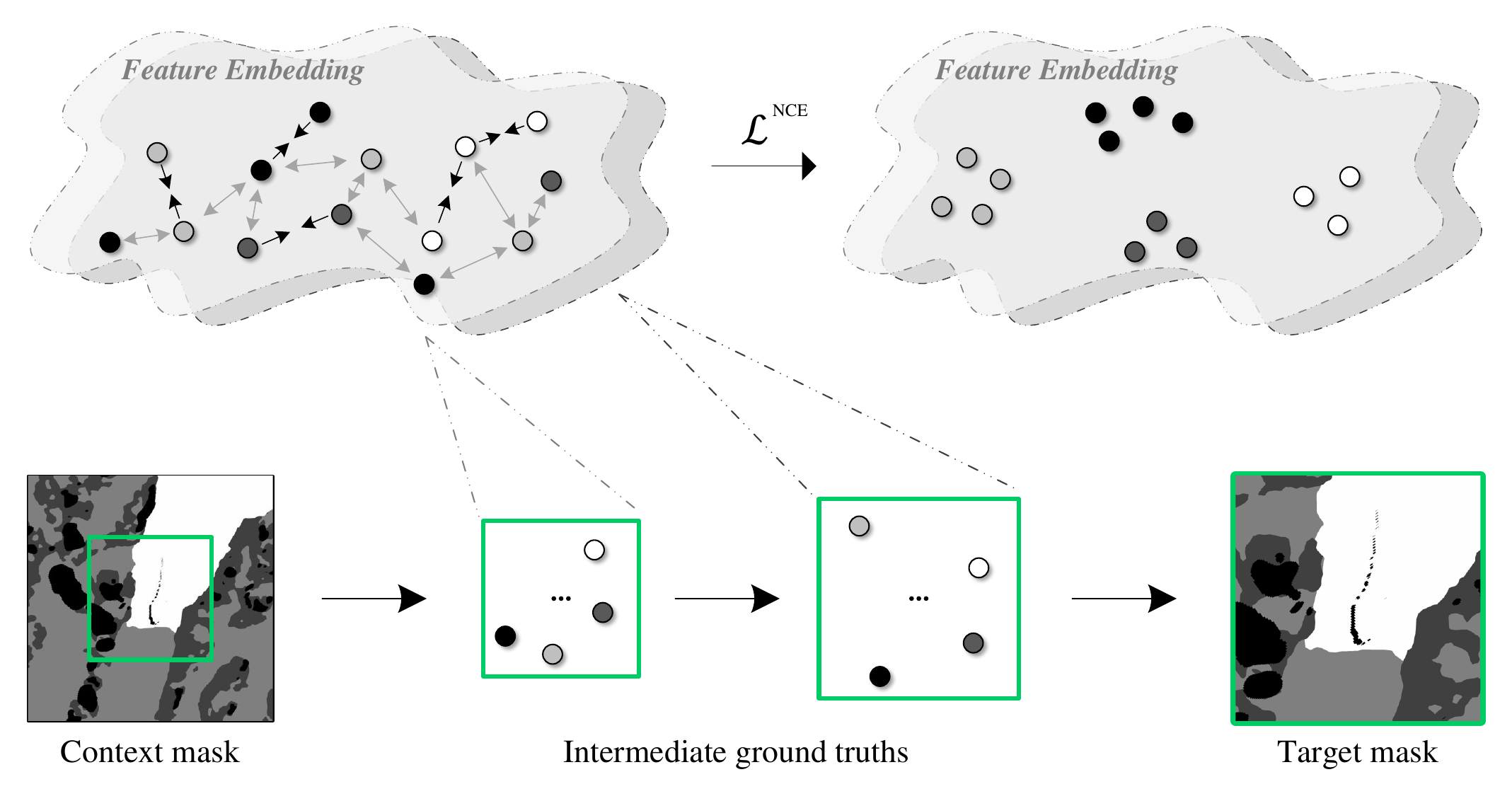}
\caption{Schematic representation of our developed pixel-to-pixel contrastive deep supervision.}\label{fig04}
\end{figure}

Inspired by these two works~\cite{10.1007/978-3-031-19809-0_1,10443562}, here we develop a pixel-to-pixel contrastive deep supervision for optimizing our hybrid model, as shown in \cref{fig04}. It integrates pixel-wise metric learning into glacier segmentation by guiding hierarchical pyramid-based pixel embeddings with category-discriminative capability. Concretely, the feature pyramid of the two ESCA-based \emph{hooking} mechanisms is supervised by the pixel-wise contrastive loss with the respective downsampled target ground truths. The pixel-wise contrastive loss is defined as:
\begin{equation}
\!\!\!\!\mathcal{L}^{\text{NCE}}_i\!=\!\frac{1}{|\mathcal{P}_i|}\!\!\sum_{\bm{i}^+\in\mathcal{P}_i\!\!}\!\!\!-_{\!}\log\frac{\exp(\bm{i}\!\cdot\!\bm{i}^{+\!\!}/\tau)}{\exp(\bm{i}\!\cdot\!\bm{i}^{+\!\!}/\tau)
+\!\sum\nolimits_{\bm{i}^{-\!}\in\mathcal{N}_i\!}\!\exp(\bm{i}\!\cdot\!\bm{i}^{-\!\!}/\tau)}
\end{equation}
where $\mathcal{P}_i$ and $\mathcal{N}_i$ represent the sets of pixel embeddings for the positive and negative samples of pixel $i$, respectively. The operator [$\cdot$] represents the dot product, and $\tau\!>\!0$ is a temperature hyper-parameter. Note that all embeddings in this loss function are normalized using $\ell_2$-normalization. The pixel-wise contrastive loss is formulated to learn an embedding space by pulling pixel samples of the same class closer together while pushing apart those from different classes. As a result, the developed pixel-to-pixel contrastive deep supervision is integrated as an auxiliary term into the overall loss function to enable joint network optimization. The total loss comprises the cross-entropy loss (CE), the dice loss (Dice), and the pixel-to-pixel contrastive deep supervision (CDS), which collectively facilitate the training of our hybrid CNN-Transformer architecture. The overall loss function, denoted as $\mathcal{L}$, is formally defined as follows: 
\begin{align}
    \mathcal{L}_{\phantom{c}} &=\lambda{_1}\mathcal{L}_{t}+\lambda{_2}\mathcal{L}_{c}+\lambda{_3}\mathcal{L}_{cds}\\
    \mathcal{L}_{t}&=\mathrm{CE}(p_\mathrm{t},y_\mathrm{t})+\mathrm{Dice}(p_\mathrm{t},y_\mathrm{t})\\
    \mathcal{L}_{c}&=\mathrm{CE}(p_\mathrm{c},y_\mathrm{c})+\mathrm{Dice}(p_\mathrm{c},y_\mathrm{c})
\end{align}
\begin{equation}
    \mathcal{L}_{cds}=\sum_{\mathrm{D=1}}^{2}\bigl(\sum\nolimits_i\mathcal{L}^{\text{NCE}}_i(\mathrm{ESCA}\mbox{-}\mathrm{hooking}{^{\mathrm{D}}_{up}},y_\mathrm{t}^{\mathrm{D}})\bigr)
\end{equation}
where $\mathcal{L}_{t}$ and $\mathcal{L}_{c}$ denote the loss functions for the target branch ($\mathrm{t}$) and the context branch ($\mathrm{c}$), respectively. $p_{*}$ represents the predictions generated by the final pixel-wise classifier while $y_{*}$ indicates the corresponding ground truth labels for each branch. $\mathcal{L}_{cds}$ refers to the pixel-to-pixel contrastive deep supervision applied to the upsampled ($*_{up}$) feature maps derived from $\mathrm{ESCA\mbox{-}hooking}^{\mathrm{D}}$, along with their respective target ground truths $y_\mathrm{t}^{\mathrm{D}}$ at the decoder depth $\mathrm{D}\in{\{1,2\}}$. The hyper-parameters $\lambda{_1},\lambda{_2},\lambda{_3}$ modulate the relative weighting of these loss components within the overall objective function.

\section{Evaluation}\label{sec:evaluation}
This section presents a thorough experimental evaluation and analysis to assess the performance of the proposed method. We commence by introducing the benchmark dataset~\cite{essd-14-4287-2022} in \cref{Dataset}, followed by a detailed exposition of the evaluation metrics and implementation details in \cref{sec:metrics} and \cref{sec:id}, respectively. In \cref{Results}, we provide a comprehensive assessment of our method, including both quantitative analyses and qualitative observations, as well as performance comparisons against current state-of-the-art methods on the aforementioned benchmark dataset~\cite{essd-14-4287-2022}. The ablation study is demonstrated in \cref{Ablation} to validate the impact of the suggested components on the overall performance. While \cref{sec:limitation} finally discusses the limitations and our future works.
\begin{table*}[!t]
	\centering
	\caption{Summary of the CaFFe dataset~\cite{essd-14-4287-2022}, including seven glacier sites with the number of images, data partition and image coverage.}
	\begin{tabular*}{\textwidth}{@{\extracolsep{\fill}}lccccccc@{\extracolsep{\fill}}}
	\toprule
    &Greenland&\multicolumn{5}{c}{Antarctic Peninsula}&Alaska \\
    \cmidrule(lr){2-2}\cmidrule(lr){3-7}\cmidrule(lr){8-8}
    &Jakobshavn Isbrae&Crane&Jorum&DBE&Sjögren-Inlet&Mapple&Columbia \\
    \cmidrule(lr){2-6}\cmidrule(lr){7-8}
    &\multicolumn{5}{c}{Training set: 559}&\multicolumn{2}{c}{Testing set: 122} \\
    \midrule
    Images [n]&159&69&77&133&121&57&65 \\
    Area [km]&\numproduct{16x19}&\numproduct{19x25}&\numproduct{20x13}&\numproduct{22x20}&\numproduct{23x19}&\numproduct{8x8}&\numproduct{32x15} \\
    \bottomrule
    \end{tabular*}
\label{tab10}
\end{table*}

\subsection{Dataset}
\label{Dataset}
CaFFe~\cite{essd-14-4287-2022} is a publicly accessible benchmark dataset for the challenging glacier calving front segmentation. It comprises 681 Synthetic Aperture Radar (SAR) images, each meticulously annotated at pixel-level through manual processes. These images were acquired from diverse geographical regions including Antarctica Peninsula, Greenland, and Alaska, which sampled by six different SAR satellite sensors (ERS, ENVISAT, RADARSAT-1, ALOS Phased Array L-band Synthetic Aperture Radar (PALSAR), TerraSAR-X (TSX) and TanDEM-X (TDX), and Sentinel-1). The dataset is partitioned into two fixed subsets: a training set consisting of 559 images and a testing set consisting of 122 images. \Cref{tab10} presents the seven glacier sites with the number of images, data split and the covered areas. Additionally, CaFFe~\cite{essd-14-4287-2022} offers two types of annotated labels per SAR image. The first type, referred to as “front” labels, delineates the calving front of marine-terminating glaciers with a 1-pixel-wide line. The second type, referred to as “zones” labels, categorizes the image into four distinct classes: ocean and ice-melange, rock outcrop, glacier, and a no-information-available class (NA-Area). To facilitate the utilization of these annotations, CaFFe~\cite{essd-14-4287-2022} introduces two baseline models which both based on the U-Net architecture~\cite{ronneberger2015u} enhanced with Atrous Spatial Pyramid Pooling (ASPP)~\cite{7913730}. Nonetheless, the “zones” model demonstrates superior performance over the “front” model with a margin of \SI{15.1}{\percent}, primarily due to the “front” model's hindered convergence and optimization caused by its pronounced class imbalance. Consequently, the “zones” labeling scheme and its corresponding baseline model are selected for subsequent experimental evaluation and comparative analysis in this study.

\subsection{Evaluation Metrics}\label{sec:metrics}
Constrained by GPU memory capacity, the full-resolution images from CaFFe~\cite{essd-14-4287-2022} dataset are partitioned into image patches via a sliding window technique. Therefore, the resulting patch-level predictions must be merged prior to performance evaluation. For this purpose, CaFFe~\cite{essd-14-4287-2022} already provides a standardized post-processing tool\footnote{\url{https://github.com/Nora-Go/Calving_Fronts_and_Where_to_Find_Them}\label{f0}}, enabling systematic assessment and analysis of the complete “zones” segmentation predictions. Glacier segmentation performance is quantified using widely adopted metrics, including precision, recall, F1-score, and intersection over union (IoU), which are formally defined as follows:
\begin{align}
\text{Precision}&=\mathrm{\frac{TP}{TP+FP}} \label{eq:pre}\\ 
\text{Recall}&=\mathrm{\frac{TP}{TP+FN}} \label{eq:rec}\\
\text{F1\mbox{-}score}&=2\times\mathrm{\frac{Precision\times Recall}{Precision+Recall}} \label{eq:f1}\\
\text{IoU}&=\mathrm{\frac{TP}{TP+FP+FN}} \label{eq:iou}
\end{align}
where $\mathrm{TP}$, $\mathrm{FP}$, $\mathrm{TN}$, and $\mathrm{FN}$ indicate the number of true positive, false positive, true negative, and false negative pixels, respectively.
\begin{figure}[!t]
\centering
\includegraphics[width=.4\textwidth]{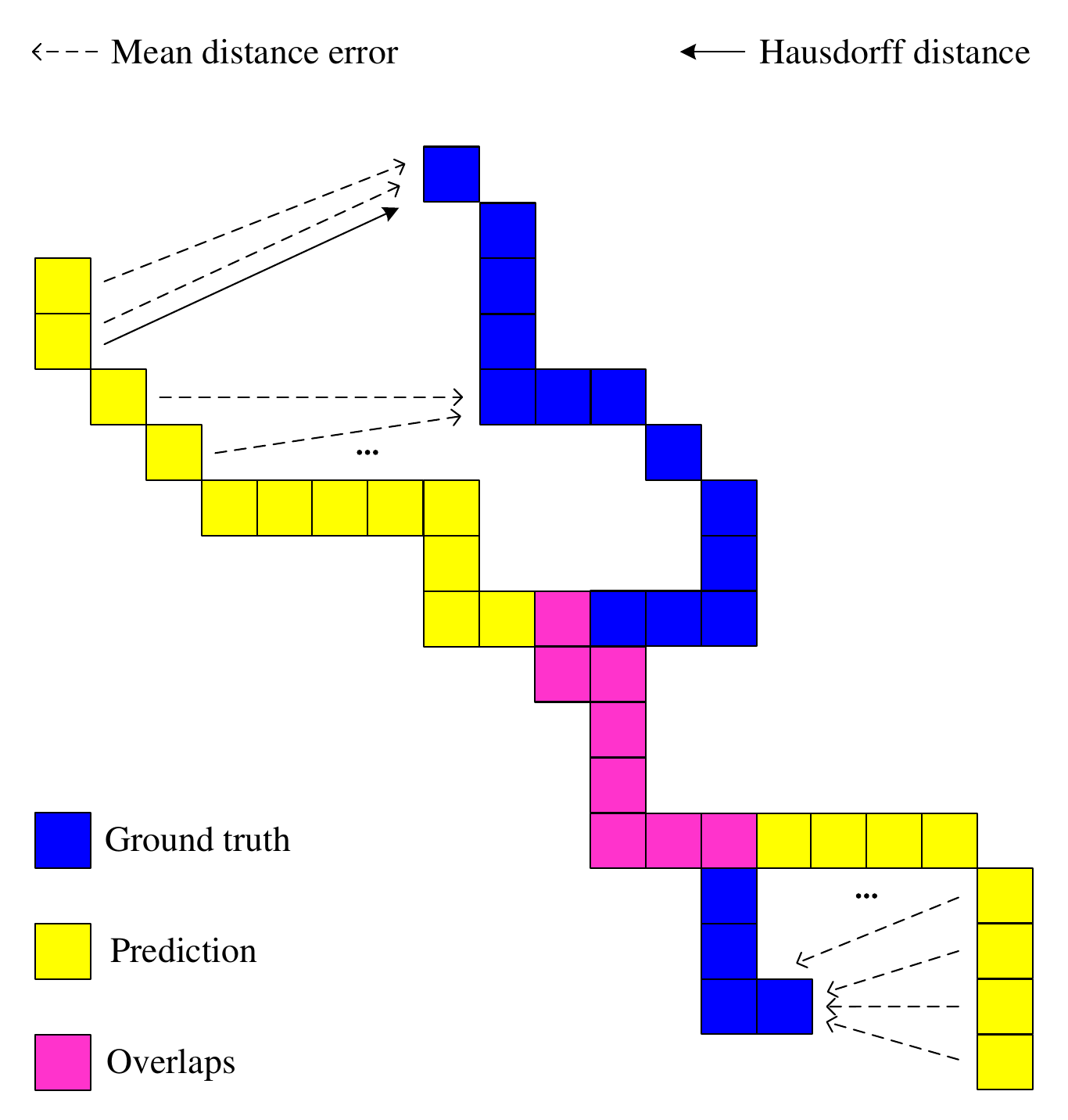}
\caption{Visualization of the two glacier calving front delineation metrics in pixel-level. The mean distance error (MDE) measures the average distance between corresponding points while the Hausdorff distance (HD) measures the maximum of the minimum distances between the two sets. Only one direction (prediction to ground truth) is depicted for better clarity.}
\label{fig05}
\end{figure}

To assess the performance of glacier calving front delineation, we follow the same post-processing tool$^{\ref{f0}}$ of the CaFFe benchmark~\cite{essd-14-4287-2022}. It employs Connected Component Analysis (CCA) on the merged “zones” segmentation predictions to enhance the ocean prediction and subsequently extracts the calving front as the boundary between the predicted ocean and glacier regions, which results in a 1-pixel-wide glacier calving front. For further details on the post-processing steps, we refer the readers to the CaFFe benchmark~\cite{essd-14-4287-2022}. The original evaluation metric for glacier calving front delineation, as established in CaFFe~\cite{essd-14-4287-2022}, is the mean distance error (MDE). It measures the average displacement in meters between the predicted calving front and its corresponding ground truth, which is defined as follows:
\begin{multline}
    \mathrm{MDE}(I)=\frac{1}{\sum_{(P,Q)\in{I}}{(|P|+|Q|)}}\ast \\ \sum_{(P,Q)\in{I}}\Bigl(\sum_{\bm{p}\in{P}}\min\limits_{\bm{q}\in{Q}}(||\bm{p}-\bm{q}||_2) + \\ \sum_{\bm{q}\in{Q}}\min\limits_{\bm{p}\in{P}}(||\bm{p}-\bm{q}||_2)\Bigr)
    \label{eqmde}
\end{multline}
where $I$ represents all the evaluated images in the test set. $P$ denotes the ground truth pixels of a particular glacier front image being tested. $Q$ refers to the predicted pixels for that same image. The notation $|*|$ denotes the cardinality, \ie the number of elements of the corresponding set. The MDE quantifies the overall positional alignment between two lines. A lower MDE indicates that, on average, the two lines are in close proximity across their entire length. However, this metric does not account for localized irregularities or extreme deviations, as exemplified by the severe jagged artifacts and other outlier points for the detected calving fronts in ViT-based approaches~\cite{10440599,cao2021swin,9994763}.

The Hausdorff distance (HD) plays a major role in any approximation method of curves, surfaces or even piecewise linear polygonal geometry~\cite{10.1007/978-3-540-79246-8_15}. It is highly sensitive to localized deviations, such as sharp spikes, dips, or regions where the curves diverge significantly. Hence, we employ the Hausdorff distance as a complementary metric to MDE, which is defined as follows:
\begin{multline}
    \mathrm{HD}(I)=\frac{1}{|I|}\ast\sum_{(P,Q)\in{I}}\max\Bigl\{\max\limits_{\bm{p}\in{P}}\min\limits_{\bm{q}\in{Q}}(||\bm{p}-\bm{q}||_2),\\\max\limits_{\bm{q}\in{Q}}\min\limits_{\bm{p}\in{P}}(||\bm{p}-\bm{q}||_2)\Bigr\}
\end{multline}

The definitions of $I$, $P$, $Q$, and $|*|$ are consistent with those provided in \cref{eqmde}. \Cref{fig05} illustrates the differences between MDE and HD during the evaluation of glacier calving front delineation at the pixel level. MDE measures the average calving front positional shift while HD evaluates detection robustness by capturing the worst-case local errors, which is crucial for analyzing complex front geometries~\cite{10.1007/978-3-540-79246-8_15}.

\subsection{Implementation Details}
\label{sec:id}
\subsubsection{Training}
We use the stochastic gradient descent (SGD) optimizer for model weights updating, configuring the momentum and weight decay with 0.9 and 1e-4, respectively. The initial learning rate is 0.01 and decays exponentially with a parameter of 0.9. The model is trained over five independent runs, each comprising 130 epochs with a batch size of 170. For the Transformer-based context branch, the patch size for ``token" extraction is configured as \numproduct{4x4} pixels, while the window size is set to \numproduct{7x7} pixels. In the \textit{Stage 1}, the hidden layers have a channel dimension ($C$) of 96. Conversely, in the CNN-based target branch, the hidden layers of the \textit{Convolutional Block 1} have a channel dimension ($C'$) of 32. Both the context and target branches process input image patches with the size of \numproduct{224x224} pixels. These two-branch inputs are online augmented with multiple random rotations within 360$^\circ$ and horizontal/vertical flips each with a probability of 0.5. There are no overlaps during the sliding window patch-extraction for the target branch. The hyper-parameters $\lambda{_1},\lambda{_2}, \lambda{_3}$ in \cref{eq7} are assigned values of 1.0, 1.0, and 0.5, respectively. ImageNet~\cite{deng2009imagenet} pre-trained model weights are used to initialize the backbone of the context branch, \ie Swin-Unet~\cite{cao2021swin} and all the Swin-Unet related model parameters in the experiment \cref{Results} and ablation study (\cref{Ablation}) to ensure a fair performance comparison. The experiments are performed on a server equipped with an AMD EPYC 7713@2.0 GHz CPU and a single Nvidia A100-SXM4 GPU. 

\begin{figure}[!t]
\centering
\includegraphics[width=.49\textwidth]{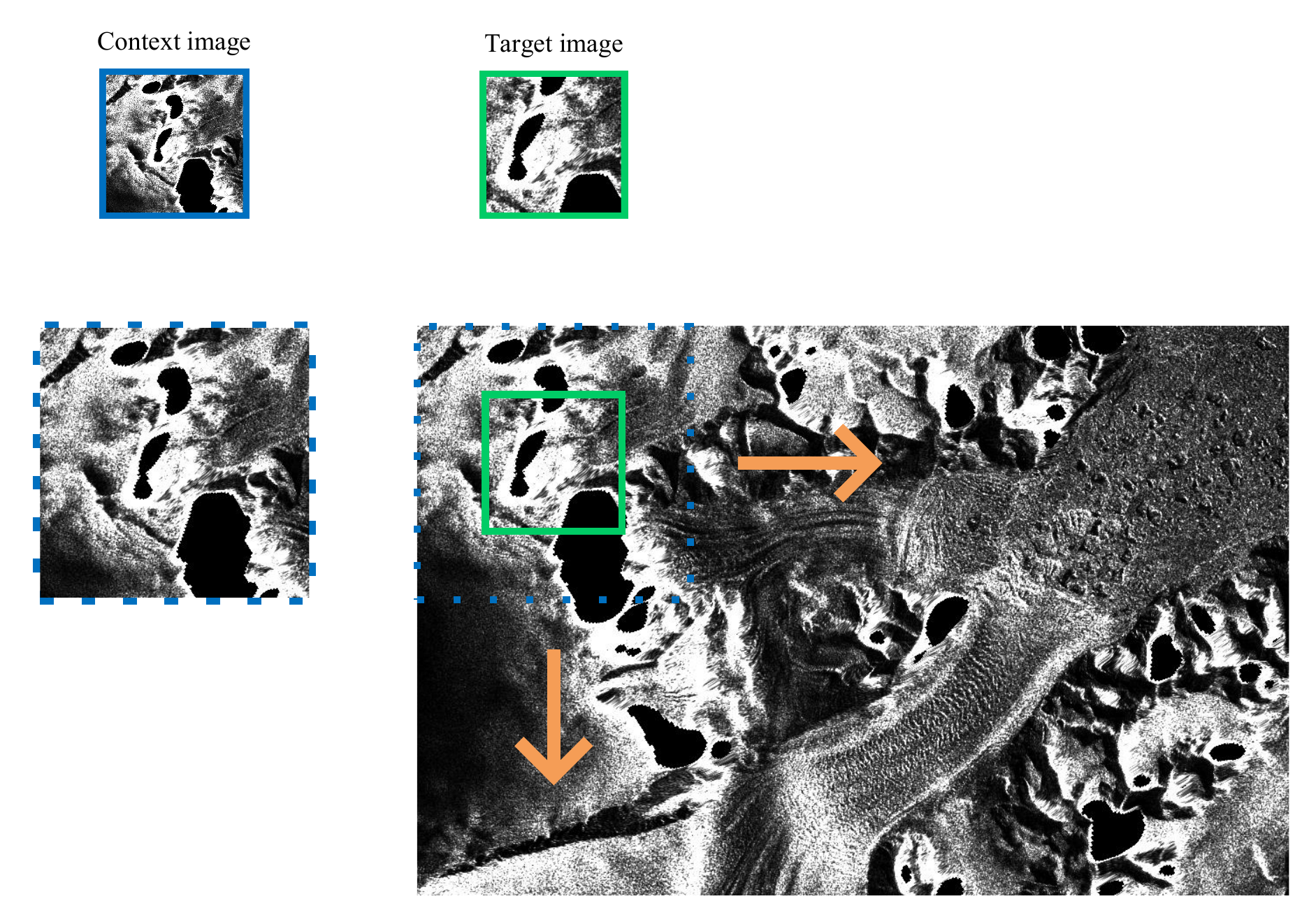}
\caption{The respective patch extraction processes for generating input images for the context branch and target branch of AMD-HookNet++.}
\label{fig13}
\end{figure}

Here we provide a schematic diagram in \cref{fig13}, illustrating the patch extraction processes for the context branch and target branch of AMD-HookNet++, respectively. 

\subsubsection{Testing}
The performance of our AMD-HookNet++ is evaluated using the aforementioned evaluation criteria on the test set of CaFFe~\cite{essd-14-4287-2022}. The experimental results of the best model determined by the validation performance within each training round is used to compute the mean and standard deviation of the evaluation metrics.
\begin{table*}[!t]
	\centering
	\caption{Benchmarking results and performance comparisons on the CaFFe dataset. The top-performing results are highlighted in \textbf{bold}, while the second-ranked results are \underline{underlined}.}
	\begin{tabular*}{\textwidth}{@{\extracolsep{\fill}}llcccccccc@{\extracolsep{\fill}}}
	\toprule
    \multirow{2}[2]{*}{Method}&\multirow{2}[2]{*}{Type}&\multirow{2}[2]{*}{Venue}&\multicolumn{4}{c}{“Zones” segmentation}&\multicolumn{3}{c}{Calving front delineation} \\
    \cmidrule(lr){4-7}\cmidrule(lr){8-10}
    &&&\bfseries{Precision}$\uparrow$&\bfseries{Recall}$\uparrow$&\bfseries{F1-score}$\uparrow$&\bfseries{IoU}$\uparrow$&\bfseries{MDE [m]}$\downarrow$&\bfseries{HD95 [m]}$\downarrow$&\bfseries{$\varnothing$}$\downarrow$ \\
	\midrule
	Baseline~\cite{essd-14-4287-2022}&CNN&\phantom{0\,\,\,\,}ESSD-2022&84.2$\pm$0.5&79.6$\pm$0.9&80.1$\pm$0.5&69.7$\pm$0.6&\phantom{0\,}753$\pm$\phantom{0}76&2,180$\pm$198&\phantom{0}1$\pm$\phantom{0}1 \\
    HookNet~\cite{van2021hooknet}&CNN&\phantom{\,\,\,\,}MedIA-2021&84.4$\pm$0.5&82.2$\pm$0.8&82.3$\pm$0.5&72.0$\pm$0.7&\phantom{0\,}588$\pm$\phantom{0}33&\mbox{-}&\phantom{0}3$\pm$\phantom{0}2 \\
    Bayesian-U-Net~\cite{9554292}&CNN&IGARSS-2021&76.4$\pm$2.3&70.7$\pm$5.8&69.8$\pm$6.3&58.0$\pm$7.2&1,011$\pm$\phantom{0}46&2,488$\pm$110&12$\pm$10 \\
    HED-U-Net~\cite{heidler2021hed}&CNN&\phantom{0\,\,\,}TGRS-2021&81.6$\pm$1.3&81.0$\pm$1.1&79.6$\pm$1.0&69.6$\pm$1.2&\phantom{0\,}646$\pm$\phantom{0}67&1,999$\pm$190&\phantom{0}6$\pm$\phantom{0}5 \\
    FineTuned-U-Net~\cite{periyasamy2022get}&CNN&\phantom{\,}JSTARS-2022&74.6$\pm$0.5&70.2$\pm$0.5&69.2$\pm$0.5&57.3$\pm$0.3&1,065$\pm$\phantom{0}47&2,741$\pm$172&12$\pm$\phantom{0}4 \\
    MTL-nnU-Net~\cite{tc-2023-34}&CNN&\phantom{0000\,\,}TC-2023&\underline{87.0$\pm$0.2}&79.1$\pm$1.7&80.7$\pm$0.1&70.8$\pm$1.8&\phantom{0\,}541$\pm$\phantom{0}84&1,952$\pm$183&\phantom{0}3$\pm$\phantom{0}1 \\
    AutoTerm~\cite{tc-17-3485-2023}&CNN&\phantom{0000\,\,}TC-2023&77.8$\pm$1.5&75.4$\pm$2.5&74.0$\pm$2.0&62.8$\pm$2.0&\phantom{0\,}909$\pm$180&2,524$\pm$418&12$\pm$\phantom{0}5 \\
    AMD-HookNet~\cite{AMD-HookNet}&CNN&\phantom{0\,\,\,}TGRS-2023&85.0$\pm$0.6&85.0$\pm$0.7&84.3$\pm$0.7&74.4$\pm$1.0&\phantom{0\,}438$\pm$\phantom{0}22&1,631$\pm$148&\phantom{0}0$\pm$\phantom{0}1 \\
    Swin-Unet~\cite{cao2021swin}&Transformer&ECCVW-2022&82.0$\pm$0.4&80.3$\pm$1.0&80.1$\pm$0.7&68.9$\pm$0.7&\phantom{0\,}576$\pm$\phantom{0}54&1,816$\pm$133&\phantom{0}4$\pm$\phantom{0}3  \\
    MISSFormer~\cite{9994763}&Transformer&\phantom{000\,}TMI-2022&78.9$\pm$0.5&75.4$\pm$0.6&75.5$\pm$0.4&64.4$\pm$0.4&1,084$\pm$\phantom{0}99&2,522$\pm$118&\phantom{0}9$\pm$\phantom{0}4  \\
    HookFormer~\cite{10440599}&Transformer&\phantom{0\,\,\,}TGRS-2024&85.9$\pm$0.3&\underline{85.5$\pm$0.4}&\underline{84.8$\pm$0.3}&\underline{75.5$\pm$0.3}&\phantom{0\,}\bfseries{353$\pm$\phantom{0}16}&\underline{1,370$\pm$\phantom{0}75}&\bfseries{\phantom{0}0$\pm$\phantom{0}0}  \\
    Trans-Unet~\cite{chen2024transunet}&Hybrid&\phantom{\,\,\,\,}MedIA-2024&80.3$\pm$0.5&78.6$\pm$1.6&77.2$\pm$1.5&66.0$\pm$0.2&\phantom{0\,}574$\pm$\phantom{0}40&1,836$\pm$198&\phantom{0}0$\pm$\phantom{0}1  \\
    AMD-HookNet++&Hybrid&\mbox{-}&\bfseries{87.9$\pm$0.2}&\bfseries{86.3$\pm$0.5}&\bfseries{86.3$\pm$0.3}&\bfseries{78.2$\pm$0.4}&\phantom{0\,}\underline{367$\pm$\phantom{0}30}&\bfseries{1,318$\pm$115}&\bfseries{\phantom{0}0$\pm$\phantom{0}0}  \\
	\bottomrule
	\end{tabular*}
\label{tab01}
\end{table*}

\subsection{Results}
\label{Results}
We first give a systematic summary of state-of-the-art approaches in \cref{tab01}, offering a comprehensive performance comparison across three type of architectures: CNN~\cite{essd-14-4287-2022,AMD-HookNet,tc-2023-34,van2021hooknet,9554292,heidler2021hed,periyasamy2022get,tc-17-3485-2023}, Transformer~\cite{cao2021swin,9994763,10440599}, and the hybrid CNN-Transformer~\cite{chen2024transunet}. Concretely, the evaluation encompasses HookNet~\cite{van2021hooknet}, Bayesian-U-Net~\cite{9554292}, HED-U-Net~\cite{heidler2021hed}, FineTuned-U-Net~\cite{periyasamy2022get}, AutoTerm~\cite{tc-17-3485-2023}, Swin-Unet~\cite{cao2021swin}, MISSFormer~\cite{9994763}, Trans-Unet~\cite{chen2024transunet}, the multi-task-learning-based MTL-nnU-Net~\cite{tc-2023-34}, the benchmark baseline: CaFFe~\cite{essd-14-4287-2022}, the base model: AMD-HookNet~\cite{AMD-HookNet}, the state-of-the-art model: HookFormer~\cite{10440599}, as well as our proposed AMD-HookNet++ within the CaFFe dataset~\cite{essd-14-4287-2022,gourmelon2025comparison}. The results demonstrate that AMD-HookNet++ exhibits superior performance compared to these methods across all evaluated metrics, with the exception of MDE. Notably, it achieves an IoU of 78.2 in glacier segmentation, surpassing the benchmark baseline: CaFFe~\cite{essd-14-4287-2022},  the base model: AMD-HookNet~\cite{AMD-HookNet}, and the state-of-the-art model: HookFormer~\cite{10440599} by significant margins of \SI{8.5}{\percent}, \SI{3.8}{\percent}, and \SI{2.7}{\percent}, respectively. Additionally, AMD-HookNet++ delivers strong performance in glacier calving front delineation with a MDE of 367\,m and a HD95 of 1,318\,m, which is on par with the state-of-the-art model: HookFormer~\cite{10440599} while outperforms the benchmark baseline: CaFFe~\cite{essd-14-4287-2022} and the base model: AMD-HookNet~\cite{AMD-HookNet} by the following absolute gains of \SI{51.3}{\percent} and \SI{16.2}{\percent} on MDE, and \SI{39.5}{\percent} and \SI{19.2}{\percent} on HD95, respectively.
\begin{table*}[!t]
	\centering
	\caption{Performance comparisons in glacier segmentation between HookFormer~\cite{10440599} and AMD-HookNet++. The top-performing results are highlighted in \textbf{bold}.}
	\begin{tabular*}{\textwidth}{@{\extracolsep{\fill}}lccccc@{\extracolsep{\fill}}}
	\toprule
    Scope&Method&\bfseries{Precision}$\uparrow$&\bfseries{Recall}$\uparrow$&\bfseries{F1-score}$\uparrow$&\bfseries{IoU}$\uparrow$\\
	\midrule
	\multirow{2}*{All}&HookFormer&85.9$\pm$0.3&85.5$\pm$0.4&84.8$\pm$0.3&75.5$\pm$0.3 \\
	&Ours&\bfseries{87.9$\pm$0.2}&\bfseries{86.3$\pm$0.5}&\bfseries{86.3$\pm$0.3}&\bfseries{78.2$\pm$0.4} \\
	\multirow{2}*{NA Area}&HookFormer&93.5$\pm$0.4&94.2$\pm$0.6&93.8$\pm$0.3&89.1$\pm$0.6 \\
	&Ours&\bfseries{99.5$\pm$0.3}&\bfseries{97.9$\pm$0.5}&\bfseries{98.7$\pm$0.1}&\bfseries{97.5$\pm$0.3} \\
	\multirow{2}*{Rock Outcrop}&HookFormer&\bfseries{81.3$\pm$0.7}&\bfseries{65.9$\pm$2.0}&\bfseries{71.8$\pm$1.6}&\bfseries{57.7$\pm$1.7} \\
	&Ours&81.3$\pm$1.3&65.9$\pm$2.9&71.6$\pm$1.2&57.6$\pm$1.4 \\
	\multirow{2}*{Glacier}&HookFormer&\bfseries{79.7$\pm$0.8}&88.4$\pm$0.7&\bfseries{83.6$\pm$0.2}&72.3$\pm$0.4 \\
	&Ours&79.3$\pm$1.3&\bfseries{88.9$\pm$1.6}&83.4$\pm$0.2&\bfseries{72.3$\pm$0.2} \\
	\multirow{2}*{Ocean and Ice Melange}&HookFormer&88.5$\pm$1.7&92.2$\pm$1.8&89.9$\pm$0.9&82.8$\pm$1.1 \\
	&Ours&\bfseries{91.6$\pm$0.9}&\bfseries{92.8$\pm$0.9}&\bfseries{91.8$\pm$0.3}&\bfseries{85.5$\pm$0.4} \\
	\bottomrule
	\end{tabular*} 
\label{tab02}
\end{table*}
\begin{figure*}[!t]
    \centering
    \includegraphics[width=\textwidth]{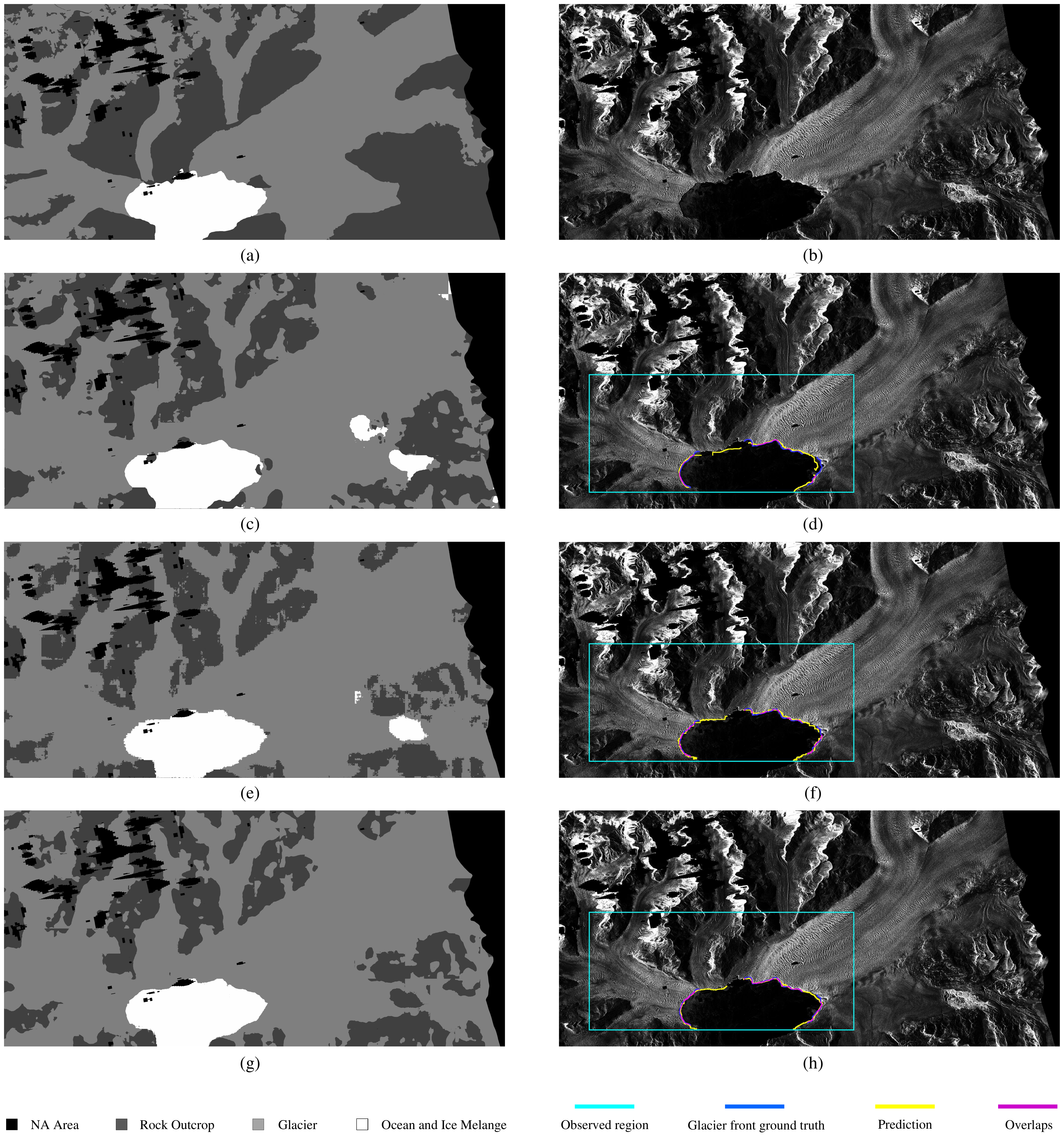}
    \caption{Qualitative comparisons of glacier segmentation on a SAR image (b) of the Columbia Glacier acquired by the TanDEM-X (TDX) satellite on 13 December 2012. (a): the “zones” segmentation ground truth of~(b). (c), (e), and (g): the predicted segmentation maps from AMD-HookNet~\cite{AMD-HookNet}, HookFormer~\cite{10440599}, and our proposed AMD-HookNet++. (d), (f), and (h): the corresponding glacier calving fronts extracted via the post-processing using the CaFFe benchmark.}
\label{fig06}
\end{figure*}
\begin{figure*}[!t]
  \centering
  \includegraphics[width=\textwidth]{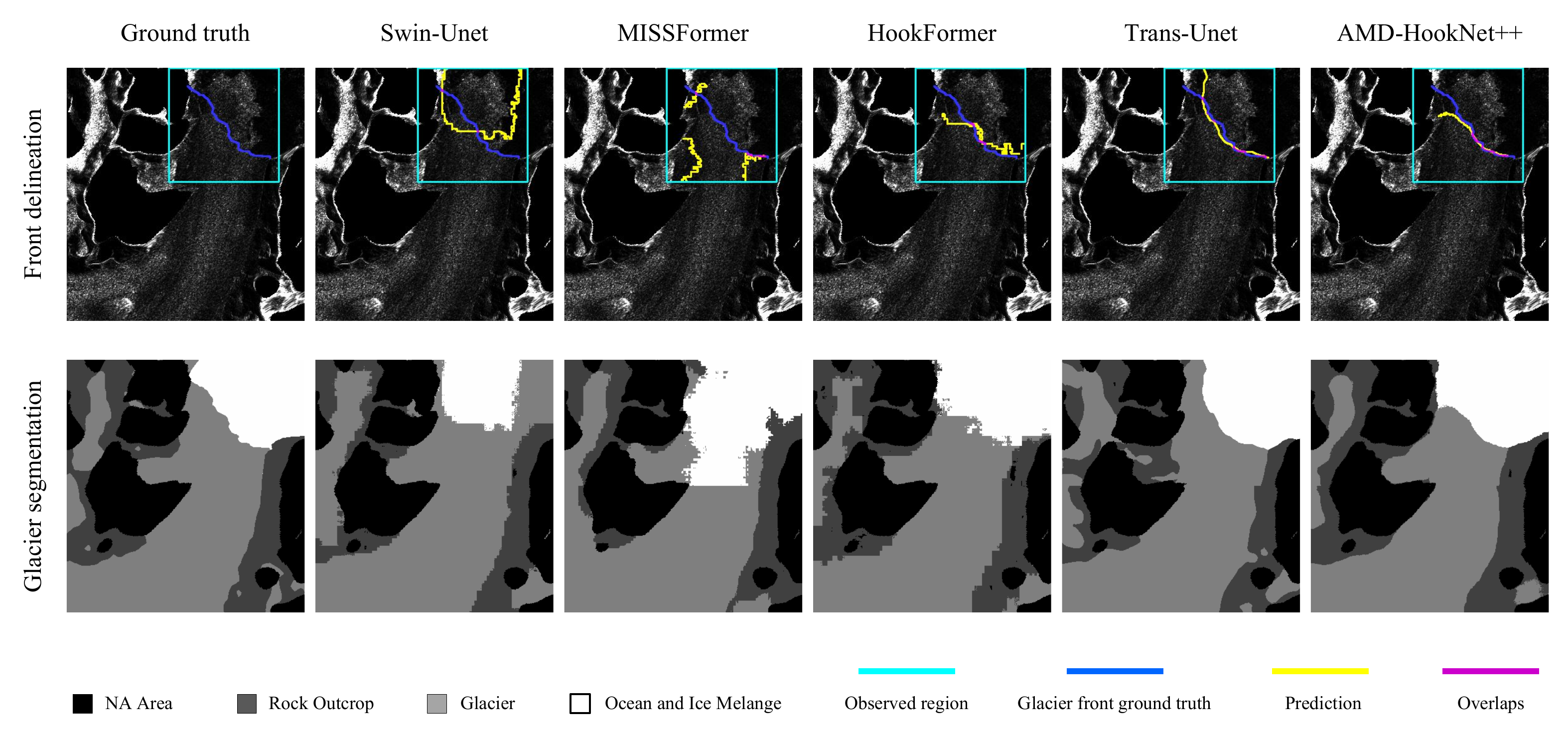}
  \caption{Qualitative comparisons of glacier segmentation and calving front delineation from Swin-Unet~\cite{cao2021swin}, MISSFormer~\cite{9994763}, HookFormer~\cite{10440599}, Trans-Unet~\cite{chen2024transunet}, and our AMD-HookNet++. The visualizations are performed on a SAR image of Mapple Glacier acquired by the ENVISAT satellite on 14 July 2007.}
  \label{figM0}
\end{figure*}
\begin{figure*}[!hp]
    \centering
    \includegraphics[width=.93\textwidth]{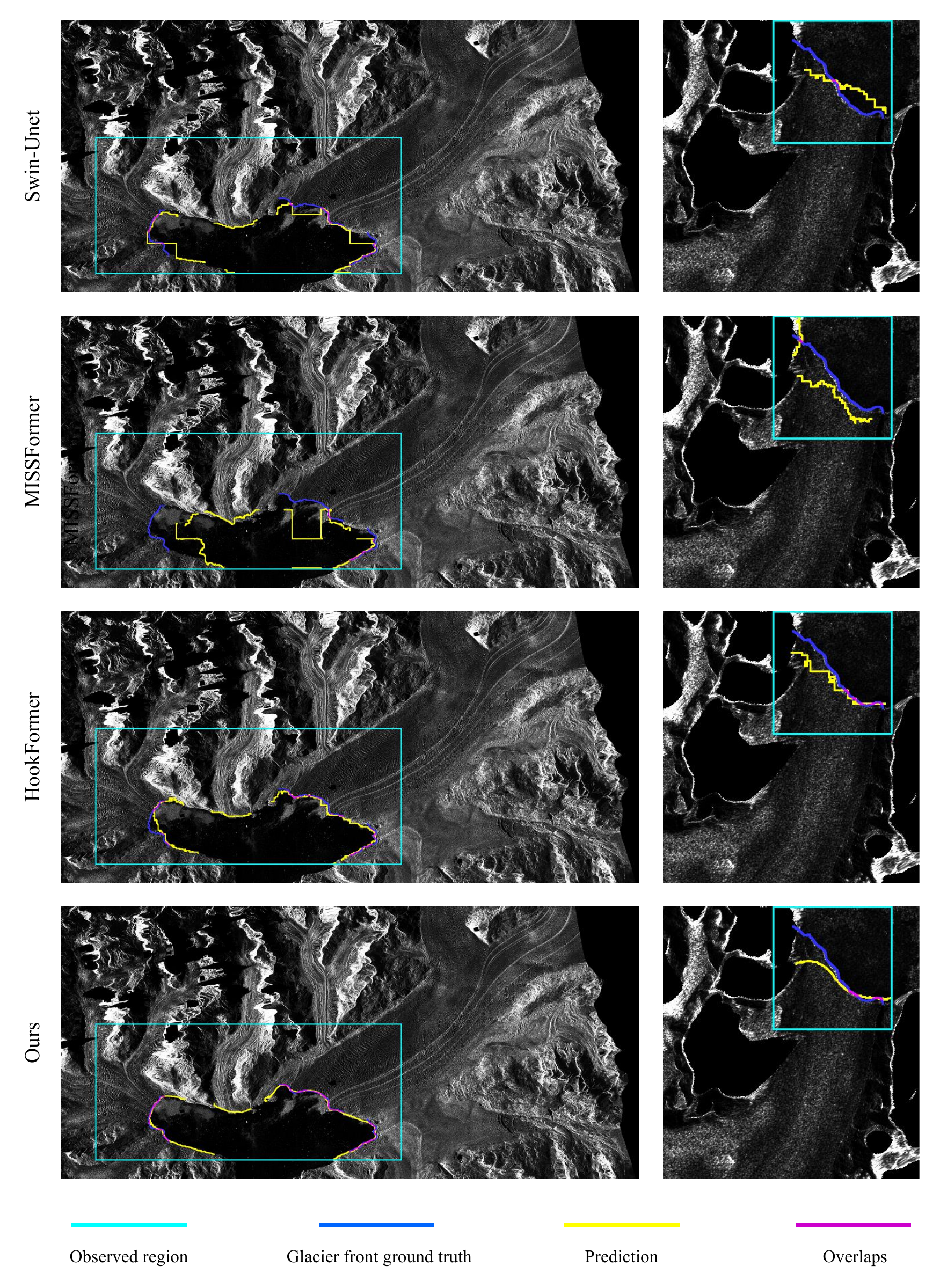}
    \caption{Qualitative comparisons of glacier calving front delineation from Swin-Unet~\cite{cao2021swin}, MISSFormer~\cite{9994763}, HookFormer~\cite{10440599}, and our AMD-HookNet++. The visualizations use two SAR images of the Columbia Glacier (left column) acquired by the TanDEM-X (TDX) satellite on 8 August 2015 and the Mapple Glacier (right column) acquired by the ENVISAT satellite on 31 March 2007.}
\label{fig07}
\end{figure*}
\begin{figure*}[!t]
    \centering
    \includegraphics[width=\textwidth]{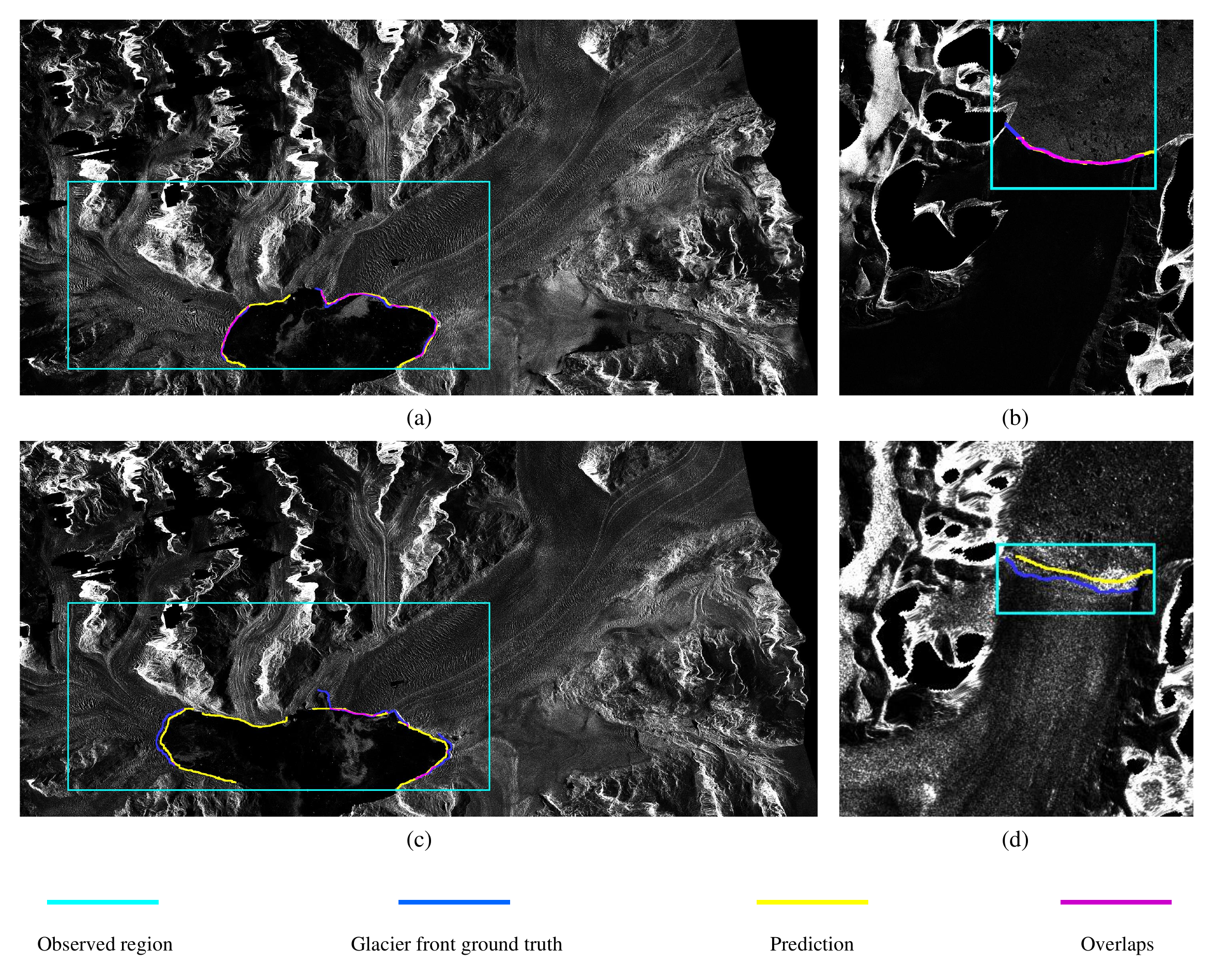}
    \caption{Accurate (top row) and inaccurate (bottom row) glacier calving front delineation visualizations of AMD-HookNet++. These SAR images are acquired by (a) the TanDEM-X (TDX) satellite on 25 August 2012 of the Columbia Glacier, (b) the TerraSAR-X (TSX) satellite on 30 December 2010 of the Mapple Glacier, (c) the TanDEM-X (TDX) satellite on 15 October 2014 of the Columbia Glacier, and (d) the Sentinel-1 satellite on the 8 September 2017 of the Mapple Glacier.}
\label{fig08}
\end{figure*}
\begin{table*}[!t]
	\centering
	\caption{Performance comparisons in glacier front delineation between HookFormer~\cite{10440599} and AMD-HookNet++, reporting the mean distance error (MDE) and the $95^\text{th}$ percentile of the Hausdorff distance (HD95) in meters. Results are categorized by glacier and season. The symbol $\varnothing$ indicates the count of instances where no fronts are predicted. The value following $\in$ specifies the number of images \wrt each case. The top-ranked results for each metric are highlighted in \textbf{bold}.}
	\begin{tabular*}{\textwidth}{@{\extracolsep{\fill}}lcccccccccc@{\extracolsep{\fill}}}
	\toprule
    \multirow{2}[2]*{Glacier}&\multirow{2}[2]*{Method}&\multirow{2}[2]*{\bfseries{MDE [m]}$\downarrow$}&\multirow{2}[2]*{\bfseries{HD95 [m]}$\downarrow$}&\multirow{2}[2]*{\bfseries{$\varnothing$}$\downarrow$}&\multicolumn{3}{c}{Summer}&\multicolumn{3}{c}{Winter}\\
    \cmidrule(lr){6-8}\cmidrule(lr){9-11}
    & & & & &\bfseries{MDE [m]}$\downarrow$&\bfseries{HD95 [m]}$\downarrow$&\bfseries{$\varnothing$}$\downarrow$&\bfseries{MDE [m]}$\downarrow$&\bfseries{HD95 [m]}$\downarrow$&\bfseries{$\varnothing$}$\downarrow$\\
	\midrule
	\multirow{2}*{All}&HookFormer&\bfseries{353$\pm$16}&1,370$\pm$\phantom{0}75&\phantom{0}0$\pm$0$\in$122&\bfseries{298$\pm$24}&\bfseries{1,078$\pm$\phantom{0}95}&0$\pm$0$\in$68&414$\pm$11&1,738$\pm$\phantom{0}83&0$\pm$0$\in$54 \\
	&Ours&367$\pm$30&\bfseries{1,318$\pm$115}&\phantom{0}0$\pm$0$\in$122&352$\pm$29&1,108$\pm$119&0$\pm$0$\in$68&\bfseries{383$\pm$33}&\bfseries{1,582$\pm$156}&0$\pm$0$\in$54 \\
	\multirow{2}*{Columbia}&HookFormer&\bfseries{386$\pm$14}&1,986$\pm$\phantom{0}85&0$\pm$0$\in$65&\bfseries{347$\pm$28}&\bfseries{1,846$\pm$\phantom{0}43}&0$\pm$0$\in$28&426$\pm$\phantom{0}9&2,092$\pm$132&0$\pm$0$\in$37 \\
	&Ours&401$\pm$31&\bfseries{1,862$\pm$\phantom{0}99}&0$\pm$0$\in$65&403$\pm$34&1,859$\pm$117&0$\pm$0$\in$28&\bfseries{398$\pm$31}&\bfseries{1,864$\pm$125}&0$\pm$0$\in$37 \\
	\multirow{2}*{Mapple}&HookFormer&\bfseries{171$\pm$27}&\phantom{0\,}\bfseries{667$\pm$131}&0$\pm$0$\in$57&\bfseries{123$\pm$25}&\phantom{0\,}\bfseries{539$\pm$145}&0$\pm$0$\in$40&278$\pm$34&\phantom{0\,}968$\pm$113&0$\pm$0$\in$17 \\
	&Ours&173$\pm$38&\phantom{0\,}698$\pm$165&0$\pm$0$\in$57&148$\pm$26&\phantom{0\,}583$\pm$141&0$\pm$0$\in$40&\bfseries{230$\pm$77}&\phantom{0\,}\bfseries{967$\pm$293}&0$\pm$0$\in$17 \\
	\bottomrule
	\end{tabular*} 
\label{tab03}
\end{table*}
\begin{table*}[!t]
	\centering
	\caption{Performance comparisons in glacier front delineation between HookFormer~\cite{10440599} and AMD-HookNet++, reporting the mean distance error (MDE) and the $95^\text{th}$ percentile of the Hausdorff distance (HD95) in meters. Results are categorized by glacier and satellite. The symbol $\varnothing$ indicates the count of instances where no fronts are predicted. The value following $\in$ specifies the number of images \wrt each case. The top-performing results are highlighted in \textbf{bold}.}
	\begin{tabular*}{\textwidth}{@{\extracolsep{\fill}}lccccccc@{\extracolsep{\fill}}}
	\toprule
    Glacier&Method&&Sentinel-1&ENVISAT&ERS&PALSAR&TSX/TDX \\
	\midrule
	\multirow{6}*{All}&\multirow{3}*{HookFormer}&\phantom{0}\bfseries{MDE [m]}$\downarrow$&1,018$\pm$\phantom{0}57&249$\pm$\phantom{0}65&195$\pm$\phantom{0}18&273$\pm$\phantom{0}90&\phantom{0\,}\bfseries{265$\pm$\phantom{0}13} \\
        &&\phantom{0}\bfseries{HD95 [m]}$\downarrow$&1,897$\pm$131&\bfseries{637$\pm$\phantom{0}91}&502$\pm$107&837$\pm$\phantom{0}64&1,311$\pm$114 \\
	&&~~\bfseries{$\varnothing$}$\downarrow$&0$\pm$0$\in$33&0$\pm$0$\in$10&0$\pm$0$\in$2&0$\pm$0$\in$8&0$\pm$0$\in$69 \\
	&\multirow{3}*{Ours}&\phantom{0}\bfseries{MDE [m]}$\downarrow$&\phantom{0\,}\bfseries{875$\pm$108}&\bfseries{231$\pm$\phantom{0}27}&\bfseries{180$\pm$\phantom{0}12}&\bfseries{226$\pm$\phantom{0}71}&\phantom{0\,}291$\pm$\phantom{0}37 \\
        &&\phantom{0}\bfseries{HD95 [m]}$\downarrow$&\bfseries{1,726$\pm$153}&728$\pm$119&\bfseries{467$\pm$\phantom{0}80}&\bfseries{707$\pm$241}&\bfseries{1,304$\pm$186} \\
	&&~~\bfseries{$\varnothing$}$\downarrow$&0$\pm$0$\in$33&0$\pm$0$\in$10&0$\pm$0$\in$2&0$\pm$0$\in$8&0$\pm$0$\in$69 \\
	\multirow{6}*{Columbia}&\multirow{3}*{HookFormer}&\phantom{0}\bfseries{MDE [m]}$\downarrow$&1,180$\pm$\phantom{0}60&$\backslash$&$\backslash$&$\backslash$&\phantom{0\,}\bfseries{280$\pm$\phantom{0}12} \\
        &&\phantom{0}\bfseries{HD95 [m]}$\downarrow$&3,202$\pm$238&$\backslash$&$\backslash$&$\backslash$&1,521$\pm$\phantom{0}41 \\
	&&~~\bfseries{$\varnothing$}$\downarrow$&0$\pm$0$\in$18&$\backslash$&$\backslash$&$\backslash$&0$\pm$0$\in$47 \\
	&\multirow{3}*{Ours}&\phantom{0}\bfseries{MDE [m]}$\downarrow$&\bfseries{1,001$\pm$129}&$\backslash$&$\backslash$&$\backslash$&\phantom{0\,}308$\pm$\phantom{0}38 \\
        &&\phantom{0}\bfseries{HD95 [m]}$\downarrow$&\bfseries{2,948$\pm$287}&$\backslash$&$\backslash$&$\backslash$&\bfseries{1,445$\pm$111} \\
	&&~~\bfseries{$\varnothing$}$\downarrow$&0$\pm$0$\in$18&$\backslash$&$\backslash$&$\backslash$&0$\pm$0$\in$47 \\
	\multirow{6}*{Mapple}&\multirow{3}*{HookFormer}&\phantom{0}\bfseries{MDE [m]}$\downarrow$&\phantom{0\,}129$\pm$\phantom{0}14&249$\pm$\phantom{0}65&195$\pm$\phantom{0}18&273$\pm$\phantom{0}90&\phantom{0\,}\bfseries{160$\pm$\phantom{0}32}  \\
        &&\phantom{0}\bfseries{HD95 [m]}$\downarrow$&\phantom{0\,}331$\pm$\phantom{00}4&\bfseries{637$\pm$\phantom{0}91}&502$\pm$107&837$\pm$\phantom{0}64&\phantom{0\,}\bfseries{863$\pm$333} \\
	&&~~\bfseries{$\varnothing$}$\downarrow$&0$\pm$0$\in$15&0$\pm$0$\in$10&0$\pm$0$\in$2&0$\pm$0$\in$8&0$\pm$0$\in$22 \\
	&\multirow{3}*{Ours}&\phantom{0}\bfseries{MDE [m]}$\downarrow$&\phantom{0\,}\bfseries{120$\pm$\phantom{0}26}&\bfseries{231$\pm$\phantom{0}27}&\bfseries{180$\pm$\phantom{0}12}&\bfseries{226$\pm$\phantom{0}71}&\phantom{0\,}167$\pm$\phantom{0}62  \\
        &&\phantom{0}\bfseries{HD95 [m]}$\downarrow$&\phantom{0\,}\bfseries{259$\pm$\phantom{0}49}&728$\pm$119&\bfseries{467$\pm$\phantom{0}80}&\bfseries{707$\pm$241}&1,000$\pm$420 \\
	&&~~\bfseries{$\varnothing$}$\downarrow$&0$\pm$0$\in$15&0$\pm$0$\in$10&0$\pm$0$\in$2&0$\pm$0$\in$8&0$\pm$0$\in$22 \\
	\bottomrule
	\end{tabular*}
\label{tab05}
\end{table*}
\begin{table*}[!t]
	\centering
	\caption{Performance comparisons in glacier front delineation between HookFormer~\cite{10440599} and AMD-HookNet++, reporting the mean distance error (MDE) and the $95^\text{th}$ percentile of the Hausdorff distance (HD95) in meters. Results are categorized by glacier and resolution. The symbol $\varnothing$ indicates the count of instances where no fronts are predicted. The value following $\in$ specifies the number of images \wrt each case. The top-performing results are highlighted in \textbf{bold}.}
	\begin{tabular*}{\textwidth}{@{\extracolsep{\fill}}lcccccccccc@{\extracolsep{\fill}}}
	\toprule
    \multirow{2}[2]*{Glacier}&\multirow{2}[2]*{Method}&\multicolumn{3}{c}{20}&\multicolumn{3}{c}{17}&\multicolumn{3}{c}{7}\\
    \cmidrule(lr){3-5}\cmidrule(lr){6-8}\cmidrule(lr){9-11}
    & &\bfseries{MDE [m]}$\downarrow$&\bfseries{HD95 [m]}$\downarrow$&\bfseries{$\varnothing$}$\downarrow$&\bfseries{MDE [m]}$\downarrow$&\bfseries{HD95 [m]}$\downarrow$&\bfseries{$\varnothing$}$\downarrow$&\bfseries{MDE [m]}$\downarrow$&\bfseries{HD95 [m]}$\downarrow$&\bfseries{$\varnothing$}$\downarrow$ \\
	\midrule
	\multirow{2}*{All}&HookFormer&\phantom{0\,}914$\pm$\phantom{0}52&1,555$\pm$101&0$\pm$0$\in$45&273$\pm$90&837$\pm$\phantom{0}64&0$\pm$0$\in$8&\bfseries{265$\pm$13}&1,311$\pm$115&0$\pm$0$\in$69 \\
	&Ours&\phantom{0\,}\bfseries{798$\pm$\phantom{0}97}&\bfseries{1,448$\pm$134}&0$\pm$0$\in$45&\bfseries{226$\pm$71}&\bfseries{707$\pm$241}&0$\pm$0$\in$8&291$\pm$37&\bfseries{1,304$\pm$186}&0$\pm$0$\in$69 \\
	\multirow{2}*{Columbia}&HookFormer&1,180$\pm$\phantom{0}60&3,203$\pm$238&0$\pm$0$\in$18&$\backslash$&$\backslash$&$\backslash$&\bfseries{280$\pm$12}&1,521$\pm$\phantom{0}41&0$\pm$0$\in$47 \\
	&Ours&\bfseries{1,001$\pm$129}&\bfseries{2,948$\pm$287}&0$\pm$0$\in$18&$\backslash$&$\backslash$&$\backslash$&308$\pm$38&\bfseries{1,445$\pm$111}&0$\pm$0$\in$47 \\
	\multirow{2}*{Mapple}&HookFormer&\phantom{0\,}182$\pm$\phantom{0}33&\phantom{0\,}457$\pm$\phantom{0}42&0$\pm$0$\in$27&273$\pm$90&837$\pm$\phantom{0}64&0$\pm$0$\in$8&\bfseries{160$\pm$32}&\phantom{0\,}\bfseries{863$\pm$333}&0$\pm$0$\in$22 \\
	&Ours&\phantom{0\,}\bfseries{169$\pm$\phantom{0}13}&\phantom{0\,}\bfseries{448$\pm$\phantom{0}35}&0$\pm$0$\in$27&\bfseries{226$\pm$71}&\bfseries{707$\pm$241}&0$\pm$0$\in$8&167$\pm$62&1,000$\pm$420&0$\pm$0$\in$22 \\
	\bottomrule
	\end{tabular*}
\label{tab06}
\end{table*}

\Cref{tab02} details the glacier segmentation performance between our AMD-HookNet++ and HookFormer~\cite{10440599}. AMD-HookNet++ obtains a precision of 87.9$\pm$0.2, a recall of 86.3$\pm$0.5, an F1-score of 86.3$\pm$0.3, and an IoU of 78.2$\pm$0.4, outperforming HookFormer~\cite{10440599} by \SI{2.0}{\percent}, \SI{0.8}{\percent}, \SI{1.5}{\percent} and \SI{2.7}{\percent}, respectively. The greatest performance gain is observed with the IoU metric, the de facto standard in semantic segmentation task. Moreover, we can see that the precise classification of the NA Area class and the Ocean and Ice Melange class dominates the enhanced performance. It is particularly important since calving fronts are located at the boundaries between the Glacier class and the Ocean and Ice Melange class. Consequently, higher accuracy in either the Glacier class or the Ocean and Ice Melange class implies improvements in both glacier segmentation and calving front detection. \Cref{fig06} presents a comparative visualization of segmentation maps, including the ground truth, AMD-HookNet~\cite{AMD-HookNet}, HookFormer~\cite{10440599}, and our proposed AMD-HookNet++, along with the corresponding calving fronts delineated using the CaFFe benchmark's post-processing pipeline~\cite{essd-14-4287-2022}. In addition, \cref{figM0} provides qualitative comparisons of glacier segmentation and calving front delineation from Swin-Unet~\cite{cao2021swin}, MISSFormer~\cite{9994763}, HookFormer~\cite{10440599}, Trans-Unet~\cite{chen2024transunet}, and our AMD-HookNet++. We can observe that all the Transformer-based approaches~\cite{cao2021swin,9994763,10440599} produce jagged edges while the hybrid models, \ie Trans-Unet~\cite{chen2024transunet} and AMD-HookNet++ demonstrate smoother calving front delineations. As previously discussed~\cite{wu2021cvt,pmlr-v238-zimerman24a,10.1007/978-3-031-19806-9_13}, the jagged edges in Transformers~\cite{cao2021swin,9994763,10440599} can be contributed to their weak inductive bias which lacks spatial locality. While the sliding window and weight-sharing mechanisms of CNNs introduce strong inductive bias (\eg the translation invariance) by operating on local neighborhood pixels, and thus enforces spatial continuity.

Following the acquisition of predicted “zones” segmentation outputs, CaFFe~\cite{essd-14-4287-2022} applies a set of post-processing operations (such as CCA) to delineate the glacier calving front. In this case, the MDE is the original metric for assessing front delineation accuracy. A comparative analysis of our AMD-HookNet++ and HookFormer~\cite{10440599} in glacier calving front delineation is presented in \cref{tab03}, with results break down by glacier and seasonal variations. AMD-HookNet++ achieves an overall MDE of 367$\pm$30\,m on all test images, which is slightly worse (\SI{4.0}{\percent}) than the MDE results of HookFormer~\cite{10440599}. However, our analysis reveals that the calving fronts extracted by HookFormer~\cite{10440599} as well as other ViT-based segmentation approaches, \eg Swin-Unet~\cite{cao2021swin} and MISSFormer~\cite{9994763} exhibit pronounced jagged artifacts on the CaFFe dataset~\cite{essd-14-4287-2022}, as shown in \cref{figM0,fig07}. These artifacts introduce substantial noise, which inevitably impedes precise observation and interpretation of glacier dynamics. Therefore, we additionally introduce the Hausdorff distance (HD) as a complementary metric to MDE, which measures the maximum degree of shape mismatch between the predicted calving fronts and the corresponding ground truths. In particular, we adopt the $95^\text{th}$ percentile of the Hausdorff distance (HD95) in this work. HD95 is widely employed as a robust metric compared to the standard HD metric, as it is less sensitive to small and extreme outliers by excluding the largest \SI{5}{\percent} of values. The results show that AMD-HookNet++ attains an overall HD95 of 1,318$\pm$115\,m across all test images, surpassing HookFormer~\cite{10440599} by \SI{3.8}{\percent} in HD95 accuracy. This highlights the superior shape similarity achieved by our hybrid CNN-Transformer model, indicating the good balance of localized spatial details in the CNN-based target branch and global relationships in the Transformer-based context branch. Moreover, AMD-HookNet++ demonstrates MDE values of 352$\pm$29\,m for summer glacier imagery and 383$\pm$33\,m for winter glacier imagery. This represents a reduced seasonal bias compared to HookFormer~\cite{10440599}, suggesting AMD-HookNet++'s improved performance in winter season which is a particularly challenging condition for extracting glacier calving fronts because of the near-identical back-scattering characteristics of glacier ice and surrounding sea ice~\cite{essd-14-4287-2022}, \ie the ice-melange. Regarding to the glacier locations, AMD-HookNet++ delivers 401$\pm$31\,m and 173$\pm$38\,m MDEs at Columbia Glacier and Mapple Glacier, respectively. In contrast, HookFormer~\cite{10440599} achieves higher MDE accuracy at Columbia Glacier (386$\pm$14\,m) and Mapple Glacier (171$\pm$27\,m). However, HookFormer~\cite{10440599} performs less accurately in HD95 accuracy at Columbia Glacier, \ie compared to the lower-resolution Mapple Glacier images (averaging \numproduct{729x688} pixels), it struggles with the high-resolution Columbia Glacier images (averaging \numproduct{1718x3650} pixels), where localized irregularities and extreme deviations occur more frequently. In this case, our AMD-HookNet++ shows its robustness by precisely capturing the full calving front geometric changes. To analyze the failure cases, two prediction examples (accurate and inaccurate) from AMD-HookNet++ are illustrated in \cref{fig08}. Unlike accurate predictions, the errors originate from two primary sources: misidentifying the coastline as the calving front, and misclassifying parts of the glacier terminus as ice-melange.

\Cref{tab05} presents a comparison of glacier calving front delineation performance between AMD-HookNet++ and HookFormer~\cite{10440599}, detailing results per glacier and satellite type. Our proposed AMD-HookNet++ demonstrates significant improvements over HookFormer in MDE accuracy, achieving absolute performance gains of \SI{14.0}{\percent}, \SI{7.2}{\percent}, \SI{7.7}{\percent}, and \SI{17.2}{\percent} on Sentinel-1, ENVISAT, ERS, and PALSAR satellites, respectively. Nevertheless, on TSX/TDX satellite, AMD-HookNet++ exhibits higher MDE prediction errors but delivers better HD95 values. Additionally, it can be observed that the MDE values of the Columbia Glacier are greater than those of the Mapple Glacier, which we associate with Columbia Glacier's more intricate geometry compared to Mapple Glacier’s simpler calving front structure~\cite{essd-14-4287-2022}.

A comparison of glacier calving front delineation performance between AMD-HookNet++ and HookFormer~\cite{10440599} across various glaciers and resolutions is shown in \cref{tab06}. The results demonstrate that AMD-HookNet++ significantly outperforms HookFormer~\cite{10440599} at \SI{20}{m} and \SI{17}{m} resolutions, with absolute improvements of \SI{12.7}{\percent} and \SI{17.2}{\percent}, respectively. In contrast, HookFormer~\cite{10440599} achieves its lowest MDE on \SI{7}{m} resolution images captured by the TSX/TDX satellite.
\begin{table*}[!t]
	\centering
	\caption{A comprehensive performance evaluation covering model architecture, “zones” segmentation, calving front delineation, model statistics, and computational efficiency (GPU memory and throughput) during the training and inference phases. The top-performing results are highlighted in \textbf{bold}.}
\begin{tabular*}{\textwidth}{@{\extracolsep{\fill}}llcccccccc@{\extracolsep{\fill}}}
	\toprule
        \multirow{3}[2]{*}{Method}&\multirow{3}[2]{*}{Type}&“Zones” segmentation&\multicolumn{3}{c}{Calving front delineation}&\multicolumn{2}{c}{Model statistics}&Training&Inference\\
        \cmidrule(lr){3-3}\cmidrule(lr){4-6}\cmidrule(lr){7-8}\cmidrule(lr){9-9}\cmidrule(lr){10-10}
        &&\multirow{2}{*}{\bfseries{IoU}$\uparrow$}&\multirow{2}{*}{\bfseries{MDE [m]}$\downarrow$}&\multirow{2}{*}{\bfseries{HD95 [m]}$\downarrow$}&\multirow{2}{*}{\bfseries{$\varnothing$}$\downarrow$}&\multirow{2}{*}{\bfseries{\#Params}$\downarrow$}&\multirow{2}{*}{\bfseries{FLOPs}$\downarrow$}&\multirow{2}{*}{\bfseries{Memory}$\downarrow$}&\bfseries{Throughput}\\
        &&&&&&&&&\bfseries{[Image/s]}$\uparrow$\\
	\midrule
    Baseline~\cite{essd-14-4287-2022} &CNN&69.7$\pm$0.6&\phantom{0\,}753$\pm$76&2,180$\pm$198&1$\pm$1&\phantom{0}19.3M&16.6G&\bfseries{\phantom{00}6.9}GB&\textbf{577.8}\\
    HED-U-Net~\cite{heidler2021hed}&CNN&69.6$\pm$1.2&\phantom{0\,}646$\pm$67&1,999$\pm$190&6$\pm$5&\bfseries{\phantom{00}8.1}M&37.9G&\phantom{0}11.5GB&195.7 \\
    AMD-HookNet~\cite{AMD-HookNet} &CNN&74.4$\pm$1.0&\phantom{0\,}438$\pm$22&1,631$\pm$148&0$\pm$1&\phantom{0}14.9M&45.7G&\phantom{0}30.2GB&182.1\\
    Swin-Unet~\cite{cao2021swin} &Transformer&68.9$\pm$0.7&\phantom{0\,}576$\pm$54&1,816$\pm$133&4$\pm$3&\phantom{0}41.4M&\bfseries{\phantom{0}8.8}G&\phantom{00}8.3GB&215.9 \\
    MISSFormer~\cite{9994763}&Transformer&64.4$\pm$0.4&1,084$\pm$99&2,522$\pm$118&9$\pm$4&\phantom{0}42.5M&\phantom{0}9.9G&\phantom{0}12.8GB&173.5 \\
    HookFormer~\cite{10440599}&Transformer&75.5$\pm$0.3&\phantom{0\,}\bfseries{353$\pm$16}&1,370$\pm$\phantom{0}75&\bfseries{0$\pm$0}&\phantom{0}59.3M&15.4G&\phantom{0}70.9GB&\phantom{0}90.9 \\
    Trans-Unet~\cite{chen2024transunet}&Hybrid&66.0$\pm$0.2&\phantom{0\,}574$\pm$40&1,836$\pm$198&0$\pm$1&105.3M&29.3G&\phantom{0}10.4GB&172.6 \\
    AMD-HookNet++&Hybrid&\bfseries{78.2$\pm$0.4}&\phantom{0\,}367$\pm$30&\bfseries{1,318$\pm$115}&\bfseries{0$\pm$0}&\phantom{0}48.6M&22.0G&\phantom{0}78.3GB&164.7 \\
	\bottomrule
\end{tabular*}
\label{tab04}
\end{table*}

Furthermore, \cref{tab04} summarizes a thorough evaluation of model performance, encompassing architectural design, “zones” segmentation, calving front delineation, model statistics, as well as computational efficiency (GPU memory and throughput) during the training and inference phases. In comparison to HookFormer~\cite{10440599}, AMD-HookNet++ achieves a parameter reduction of \SI{18.0}{\percent}. Notably, when benchmarked against Trans-Unet~\cite{chen2024transunet}, a representative hybrid CNN-Transformer architecture, AMD-HookNet++ achieves a substantial parameter reduction of \SI{47.9}{\percent} while delivering significant performance improvements with \SI{12.2}{\percent} on IoU, \SI{36.1}{\percent} on MDE, and \SI{28.2}{\percent} on HD95, respectively. While AMD-HookNet++ has higher computational complexity (FLOPs) than HookFormer~\cite{10440599}, it actually achieves a significantly faster inference speed (164.7 \vs 90.9 images/s). This underscores that FLOPs is an indirect performance metric and does not always correlate with real-world speed, as also pointed out in ShuffleNet V2~\cite{10.1007/978-3-030-01264-9_8}. This phenomenon is further exemplified by comparing Trans-Unet~\cite{chen2024transunet} to HookFormer~\cite{10440599} and AMD-HookNet++. Despite having substantially more parameters and FLOPs than either HookFormer~\cite{10440599} or AMD-HookNet++, Trans-Unet~\cite{chen2024transunet} achieves the highest inference speed of all (172.6 images/s). This confirms that architectural design is a critical determinant of actual runtime performance.
\begin{table*}[!t]
\centering
\captionsetup{position=bottom}
\caption{Ablation study evaluating different components. The top-performing results are highlighted in \textbf{bold}.}
\begin{subtable}{.99\textwidth}
\begin{NiceTabular*}{.99\textwidth}{@{\extracolsep{\fill}}cccc|cccccccccc@{\extracolsep{\fill}}}
	\toprule
        \multirow{3}[2]{*}{\#1}&\multirow{3}[2]{*}{\#2}&\multirow{3}[2]{*}{\#3}&\multirow{3}[2]{*}{\#4}&\multirow{3}[2]{*}{\bfseries{\#Params}$\downarrow$}&\multirow{3}[2]{*}{\bfseries{FLOPs}$\downarrow$}&\multicolumn{8}{c}{Pretraining}\\
        &&&&&&\multicolumn{4}{c}{ImageNet}&\multicolumn{4}{c}{Scratch}\\
        \cmidrule(lr){7-10}\cmidrule(lr){11-14}
        &&&&&&\bfseries{IoU}$\uparrow$&\bfseries{MDE [m]}$\downarrow$&\bfseries{HD95 [m]}$\downarrow$&\bfseries{$\varnothing$}$\downarrow$&\bfseries{IoU}$\uparrow$&\bfseries{MDE [m]}$\downarrow$&\bfseries{HD95 [m]}$\downarrow$&\bfseries{$\varnothing$}$\downarrow$\\
	\midrule
    \checkmark&&&&\bfseries{48.23M}&\bfseries{21.8G}&76.7$\pm$0.7&425$\pm$23&1,467$\pm$\phantom{0}79&0$\pm$1&67.5$\pm$0.3&938$\pm$51&2,589$\pm$\phantom{0}59&9$\pm$1 \\
    \checkmark&\checkmark&&&48.25M&21.9G&77.0$\pm$0.4&437$\pm$11&1,440$\pm$\phantom{0}44&0$\pm$1&68.3$\pm$0.8&\bfseries{823$\pm$20}&2,357$\pm$114&\bfseries{7$\pm$1} \\
    \checkmark&&\checkmark&&48.60M&22.0G&77.2$\pm$0.6&395$\pm$18&1,429$\pm$\phantom{0}50&\bfseries{0$\pm$0}&68.9$\pm$0.6&863$\pm$36&2,281$\pm$\phantom{0}70&8$\pm$1\\
    \checkmark&\checkmark&\checkmark&&48.63M&22.0G&77.5$\pm$0.6&404$\pm$54&1,411$\pm$\phantom{0}54&\bfseries{0$\pm$0}&69.3$\pm$0.5&854$\pm$17&2,292$\pm$\phantom{0}55&8$\pm$1\\
    \checkmark&\checkmark&\checkmark&\checkmark&\mbox{-}&\mbox{-}&\bfseries{78.2$\pm$0.4}&\bfseries{367$\pm$30}&\bfseries{1,318$\pm$115}&\bfseries{0$\pm$0}&\bfseries{69.5$\pm$0.9}&831$\pm$43&\bfseries{2,275$\pm$\phantom{0}92}&\bfseries{7$\pm$1} \\
	\bottomrule
\end{NiceTabular*}
 \caption{Proposed components ablation with/without ImageNet pretraining, where \#1 denotes the CNN-Transformer hybridization, \#2 denotes the spatial attention of the ESCA module, \#3 denotes the channel attention of the ESCA module, and \#4 denotes the contrastive deep supervision.}
 \label{tab09}
\end{subtable}
\\[1em]
\begin{subtable}{.99\textwidth}
 \begin{NiceTabular*}{.99\textwidth}{@{\extracolsep{\fill}}ccccc|cccccc@{\extracolsep{\fill}}}
	\toprule
        CDS&SA~\cite{AMD-HookNet}&SENet~\cite{8701503}&CBAM~\cite{10.1007/978-3-030-01234-2_1}&ESCA&\bfseries{\#Params}$\downarrow$&\bfseries{FLOPs}$\downarrow$&\bfseries{IoU}$\uparrow$&\bfseries{MDE [m]}$\downarrow$&\bfseries{HD95 [m]}$\downarrow$&\bfseries{$\varnothing$}$\downarrow$ \\
	\midrule
 	\checkmark&\checkmark&&&&49.40M&22.3G&77.6$\pm$0.6&414$\pm$27&1,441$\pm$\phantom{0}60&0$\pm$1 \\
        \checkmark&&\checkmark&&&\bfseries{48.29}M&\bfseries{21.9G}&77.6$\pm$0.4&397$\pm$13&1,380$\pm$\phantom{0}76&0$\pm$1 \\
        \checkmark&&&\checkmark&&48.62M&\bfseries{21.9G}&77.8$\pm$0.5&421$\pm$41&1,497$\pm$148&\bfseries{0$\pm$0} \\
        \checkmark&&&&\checkmark&48.63M&22.0G&\bfseries{78.2$\pm$0.4}&\bfseries{367$\pm$30}&\bfseries{1,318$\pm$115}&\bfseries{0$\pm$0} \\
	\bottomrule
 \end{NiceTabular*}
 \caption{Attention module ablation under the contrastive deep supervision (CDS).}
 \label{tab07}
\end{subtable}
\\[1em]
\begin{subtable}{.99\textwidth}	
 \begin{NiceTabular*}{.99\textwidth}{@{\extracolsep{\fill}}ccc|cccc@{\extracolsep{\fill}}}
	\toprule
        ESCA&DS~\cite{lee2015deeply}&CDS&\bfseries{IoU}$\uparrow$&\bfseries{MDE [m]}$\downarrow$&\bfseries{HD95 [m]}$\downarrow$&\bfseries{$\varnothing$}$\downarrow$\\
	\midrule
	\checkmark&&&77.5$\pm$0.6&404$\pm$54&1,411$\pm$\phantom{0}54&\bfseries{0$\pm$0} \\
        \checkmark&\checkmark&&77.7$\pm$0.4&381$\pm$29&1,388$\pm$\phantom{0}79&0$\pm$1 \\
        \checkmark&&\checkmark&\bfseries{78.2$\pm$0.4}&\bfseries{367$\pm$30}&\bfseries{1,318$\pm$115}&\bfseries{0$\pm$0} \\
	\bottomrule
 \end{NiceTabular*}
 \caption{Deep supervision ablation under the ESCA module.}
 \label{tab08}
\end{subtable}
\label{tab11}
\end{table*}

\subsection{Ablation Study}\label{Ablation}
In this section, we perform three ablation studies to evaluate the contribution and necessity of the proposed components: (1) the proposed components with/without ImageNet pretraining~\cite{deng2009imagenet}, (2) the enhanced spatial-channel attention (ESCA) module, and (3) the developed pixel-to-pixel contrastive deep supervision strategy. These experiments are designed to systematically assess the impact of each component on the overall model performance.

The ablation study evaluating the contribution of each proposed component, with and without ImageNet pretraining~\cite{deng2009imagenet}, is detailed in \cref{tab09}. The ablated components include:
\begin{enumerate}
    \item[\#1] The CNN-Transformer hybridization.
    \item[\#2] The spatial attention of the ESCA module.
    \item[\#3] The channel attention of the ESCA module. 
    \item[\#4] The contrastive deep supervision.
\end{enumerate}

We can see that the complete ESCA module and the contrastive deep supervision are critical drivers of the performance gains over the base hybrid architecture. Concretely, the full ESCA module improves performance by \SI{0.8}{\percent} on IoU, \SI{4.9}{\percent} on MDE, and \SI{3.8}{\percent} on HD95, respectively. The results are further enhanced by the introduction of contrastive deep supervision, yielding consistent improvements of \SI{1.5}{\percent} on IoU, \SI{13.6}{\percent} on MDE, and \SI{10.2}{\percent} on HD95, respectively. Within the ESCA module, the channel attention component demonstrates a more robust impact on performance enhancement compared to the spatial attention, which we attribute to the channel-wise feature concatenation in our hybrid fusion paradigm. Notably, all models pretrained on ImageNet~\cite{deng2009imagenet} substantially outperform the comparable baselines without it, indicating that large-scale dataset pretraining is an essential factor for the initialization of the Transformer architecture.
\begin{figure}[!t]
\centering
\includegraphics[width=.48\textwidth]{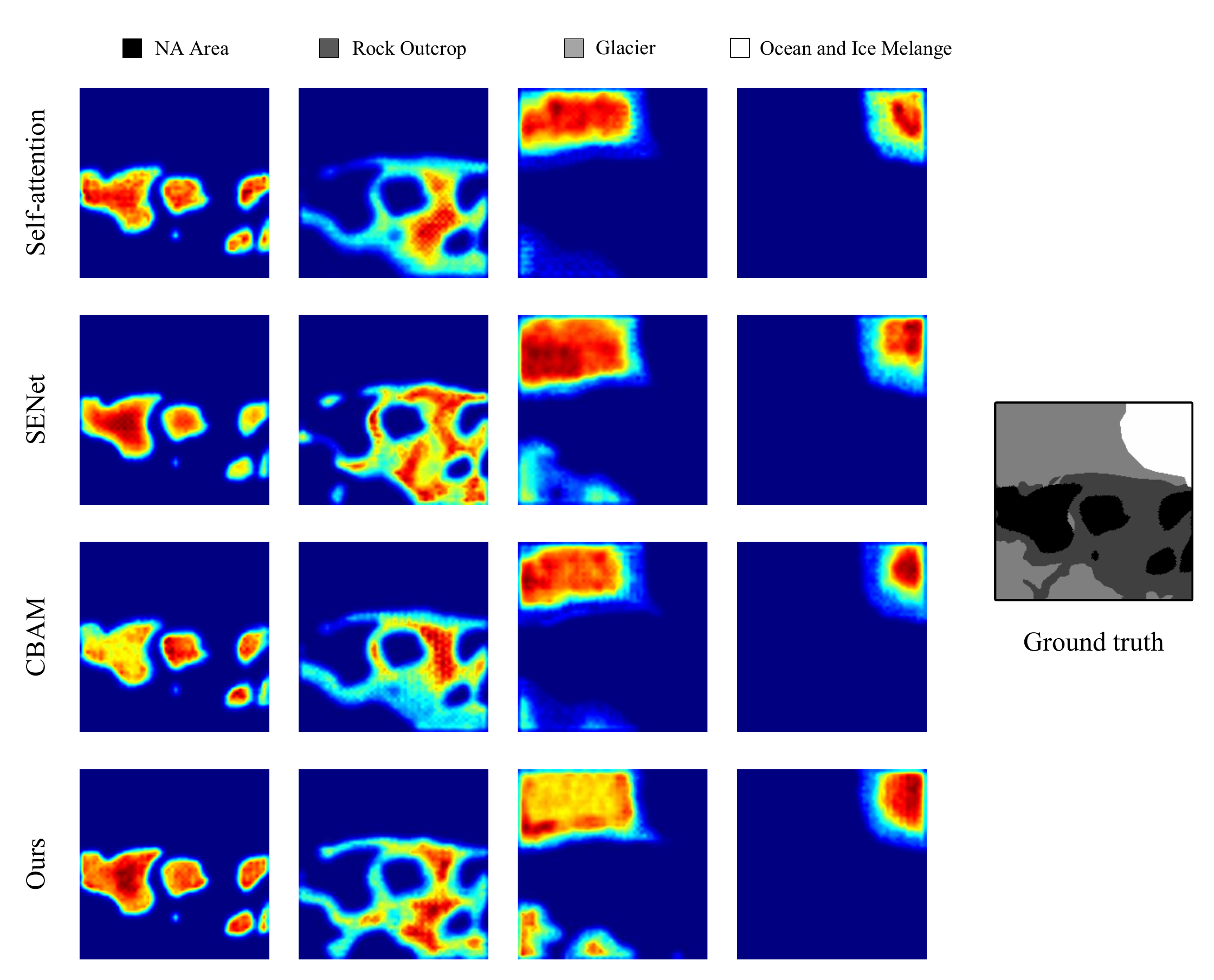}
\caption{Visualization of attention maps for the second \emph{hooking} layer of self-attention~\cite{AMD-HookNet}, SENet~\cite{8701503}, CBAM~\cite{10.1007/978-3-030-01234-2_1}, and our ESCA module. Best viewed zoomed-in.}
\label{fig12}
\end{figure}

The ablation study regarding the attention-based \emph{hooking} mechanism of AMD-HookNet++ is presented in \cref{tab07}. We evaluate the spatial self-attention (SA) block from AMD-HookNet~\cite{AMD-HookNet}, the squeeze-and-excitation block from SENet~\cite{8701503}, the convolutional block attention module from CBAM~\cite{10.1007/978-3-030-01234-2_1}, and our proposed ESCA module as part of the ablation analysis. The experiments reveal that SENet~\cite{8701503} achieves lower MDE and HD95 values than the SA block~\cite{AMD-HookNet}, suggesting that channel-wise attention is more critical than spatial attention for our channel-concatenated hybrid feature fusion. On the other hand, CBAM~\cite{10.1007/978-3-030-01234-2_1}, which incorporates both spatial and channel attention, outperforms the SA block~\cite{AMD-HookNet} and SENet~\cite{8701503} on the IoU metric, confirming that a combined attention mechanism is more effective. These findings motivate our ESCA module, which replaces static pooling operations (e.g., global average/max pooling in SENet~\cite{8701503} and CBAM~\cite{10.1007/978-3-030-01234-2_1}) with dynamic spatial-channel attention recalibration, leading to superior CNN-Transformer feature fusion with \SI{0.6}{\percent}, \SI{0.6}{\percent}, and \SI{0.4}{\percent} IoU improvement over AMD-HookNet~\cite{AMD-HookNet}, SENet~\cite{8701503}, and CBAM~\cite{10.1007/978-3-030-01234-2_1}, respectively. Moreover, we provide a comparative visualization of the attention maps for the second \emph{hooking} layer in \cref{fig12}, contrasting the self-attention~\cite{AMD-HookNet}, SENet~\cite{8701503}, CBAM~\cite{10.1007/978-3-030-01234-2_1}, and our proposed ESCA module. In this case, ESCA offers more precise attention to the fine-grained details within the Rock Outcrop class and the Glacier class.

\begin{figure}[!t]
\centering
\begin{subfigure}{.24\textwidth}
\includegraphics[width=\textwidth]{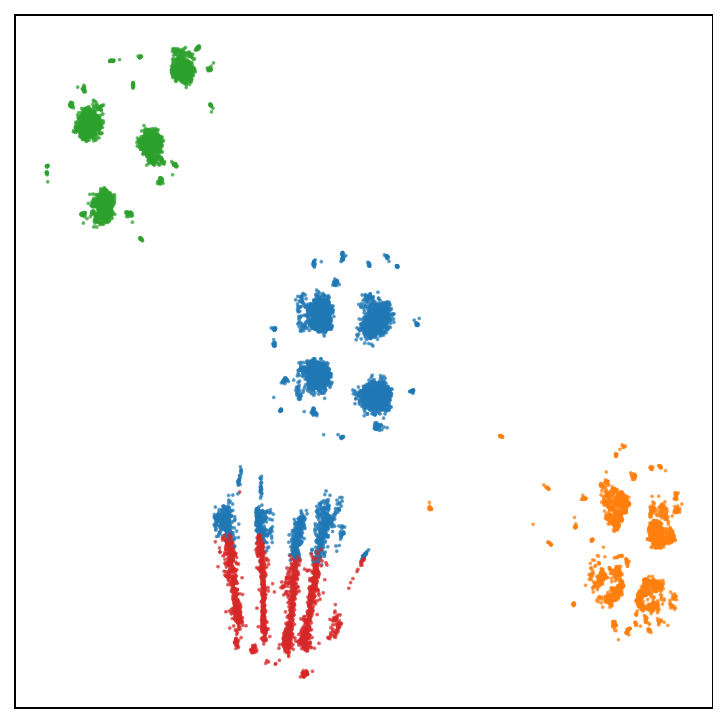}
\caption{}
\label{fig10}
\end{subfigure}
\begin{subfigure}{.24\textwidth}
\includegraphics[width=\textwidth]{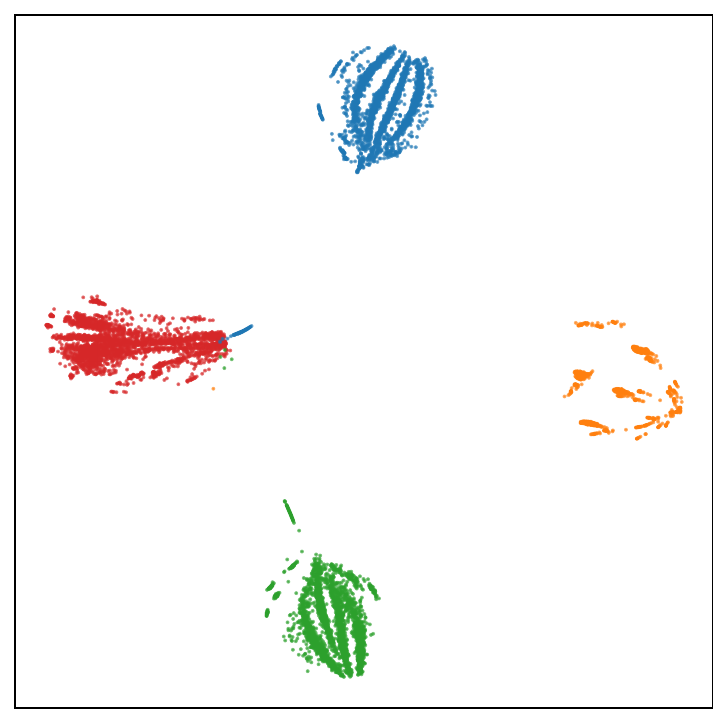}
\caption{}
\label{fig11}
\end{subfigure}
\caption{t-SNE visualizations of the feature space derived from the second \emph{hooking} position which learned with \subref{fig10} deep supervision and \subref{fig11} our developed pixel-to-pixel contrastive deep supervision. The features are color-coded based on their respective class labels.}
\label{fig09}
\end{figure}

The ablation study regarding deep supervision strategy of AMD-HookNet++ is reported in \cref{tab08}. Consistent with conclusions of AMD-HookNet~\cite{AMD-HookNet} and HookFormer~\cite{10440599}, our experiments confirm that deep supervision (DS)~\cite{lee2015deeply} stabilizes training in two-branch architectures, boosting performance by \SI{0.2}{\percent} IoU, \SI{6.0}{\percent} MDE and \SI{1.6}{\percent} HD95 compared to the model trained without it. However, the intermediate layers learn task-irrelevant features which can be improved by integrating metric learning, \ie our developed pixel-to-pixel contrastive deep supervision strategy (CDS), which achieves \SI{0.5}{\percent} IoU, \SI{3.7}{\percent} MDE, and \SI{0.5}{\percent} HD95 performance gains over regular deep supervision. \Cref{fig09} visualizes the feature space derived from the second \emph{hooking} position that is learned with \subref{fig10} deep supervision and \subref{fig11} our developed pixel-to-pixel contrastive deep supervision using t-Distributed Stochastic Neighbor Embedding (t-SNE). It can be seen that our approach produces a more structured semantic embedding space, effectively enhancing the category-discriminative power of hierarchical pyramid-based pixel embeddings.

\subsection{Limitations and Future Works}\label{sec:limitation}
The above experimental results and analyses verify the robustness and effectiveness of our proposed AMD-HookNet++. However, as evidenced by \cref{tab09}, a significant performance gap exists between models with and without ImageNet pretraining~\cite{deng2009imagenet}, underscoring the value of domain-specific knowledge for glacier segmentation. Therefore, in our future works, we will explore the integration of more advanced representation learning techniques, \eg the vision foundation models tailored for remote sensing~\cite{li2025fleximo, li2025urbansam} or self-supervised pretraining strategies~\cite{9157100} to further advance glacier segmentation, building upon the methodological innovations introduced in the current study.

\section{Conclusion}\label{sec:conclusion}
In this article, we propose AMD-HookNet++, a novel advanced hybrid CNN-Transformer feature enhancement method for glacier calving front segmentation. Considering the balance of localized spatial details in CNN and global relationships in Transformer, two branches designed with distinct purposes are involved: the Transformer-based context branch is responsible for capturing long-range dependencies and delivering  global contextual information within a larger view, which assists the CNN-based target branch in improving local spatial fine-grained details. To foster interactions between the hybrid CNN-Transformer branches, we devise an enhanced spatial-channel attention (ESCA) module to dynamically adjust the token relationships from both spatial and channel perspectives. Additionally, we develop a pixel-to-pixel contrastive deep supervision for optimizing our hybrid model. It integrates pixel-wise metric learning into glacier segmentation by guiding hierarchical pyramid-based pixel embeddings with category-discriminative capability. Through extensive experiments and comprehensive quantitative and qualitative analyses, we show that AMD-HookNet++ sets a new state-of-the-art glacier segmentation performance while maintaining competitive calving front detection accuracy on the challenging benchmark dataset CaFFe. More importantly, our hybrid model produces smoother delineations of calving fronts, resolving the issue of jagged edges typically seen in pure Transformer-based approaches.

\section*{Acknowledgments}
Grateful acknowledgment is made to the German Aerospace Center (DLR), the European Space Agency (ESA), and the Alaska Satellite Facility (ASF) for providing the SAR data used in the benchmark data set. The authors thank the Bavarian State Ministry of Science and the Arts for financial support within the International Doctorate Program “Measuring and Modelling Mountain glaciers and ice caps in a Changing ClimAte (M³OCCA)” by the Elite Network of Bavaria. The authors thank the German Research Foundation (DFG) project “Large-scale Automatic Calving Front Segmentation and Frontal Ablation Analysis of Arctic Glaciers using Synthetic-Aperture Radar Image Sequences (LASSI)” (Project number: 512625584) and “PAGE” within the DFG Emmy Noether Programme (DFG – SE3091/3-1; DFG – CH2080/5-1; DFG – SE3091/4-1). The authors thank the financial support within the framework of an ESA Living Planet Fellowship (Grant: MIT-AP). Moreover, the authors gratefully acknowledge the scientific support and High Performance Computing (HPC) resources provided by the Erlangen National High Performance Computing Center (NHR@FAU) of the Friedrich-Alexander-Universität Erlangen-Nürnberg (FAU) under the NHR project – b194dc. NHR funding is provided by federal and Bavarian state authorities. NHR@FAU hardware is partially funded by the DFG – 440719683.
        
\section*{Implementation}\label{sec:implementation}
The codes for reproducing the experimental results in this work are available on GitHub~\cite{WinNT}.

\bibliographystyle{IEEEtran}
\bibliography{References}

\end{document}